\documentclass{ieeeaccess}
\usepackage{cite}
\usepackage{amsmath,amssymb,amsfonts}
\usepackage{algorithmic}
\usepackage[dvipdfmx]{graphicx}
\usepackage{textcomp}
\usepackage{bm}
\usepackage{multirow}
\usepackage{color}
\newcommand{\ctext}[1]{\raise0.2ex\hbox{\textcircled{\scriptsize{#1}}}}

\def\BibTeX{{\rm B\kern-.05em{\sc i\kern-.025em b}\kern-.08em
    T\kern-.1667em\lower.7ex\hbox{E}\kern-.125emX}}
\begin{document}
\history{Date of publication xxxx 00, 0000, date of current version xxxx 00, 0000.}
\doi{10.1109/ACCESS.2017.DOI}

\title{Imitation Learning for Nonprehensile Manipulation through Self-Supervised Learning Considering Motion Speed}
\author{\uppercase{YUKI SAIGUSA}\authorrefmark{1}, \IEEEmembership{Non Member},
\uppercase{SHO SAKAINO}\authorrefmark{2},\IEEEmembership{Member, IEEE}, 
\\AND \uppercase{TOSHIAKI TSUJI}\authorrefmark{3}, \IEEEmembership{Senior Member, IEEE}}
\address[1]{Faculty of Engineering, Information and Systems University of Tsukuba, Ibaraki 305-8577, Japan
 (e-mail:saigusa.yuki.xp@alumni.tsukuba.ac.jp)}
\address[2]{Department of Intelligent Interactive Systems, University of Tsukuba, Ibaraki 305-8577, Japan}
\address[3]{Graduate School of Science and Engineering, Saitama University, Saitama 338-8570, Japan}
\tfootnote{This study was partly supported by the Adaptable and Seamless Technology Transfer Program through Target-driven R\&D (A-STEP) from the
Japan Science and Technology Agency (JST) Grant Number JPMJTR20RG and the Japan Society for the Promotion of Science by a Grant-in-Aid for Scientific Research (B) under Grant 21H01347. }

\markboth
{Author \headeretal: Preparation of Papers for IEEE TRANSACTIONS and JOURNALS}
{Author \headeretal: Preparation of Papers for IEEE TRANSACTIONS and JOURNALS}

\corresp{Corresponding author: Yuki Saigusa (e-mail:saigusa.yuki.xp@alumni.tsukuba.ac.jp)}

\begin{abstract} 
Robots are expected to replace menial tasks such as housework. Some of these tasks include nonprehensile manipulation performed without grasping objects. Nonprehensile manipulation is very difficult because it requires considering the dynamics of environments and objects. Therefore imitating complex behaviors requires a large number of human demonstrations. In this study, a self-supervised learning that considers dynamics to achieve variable speed for nonprehensile manipulation is proposed. The proposed method collects and fine-tunes only successful action data obtained during autonomous operations. By fine-tuning the successful data, the robot learns the dynamics among itself, its environment, and objects.
We experimented with the task of scooping and transporting pancakes using the neural network model trained on 24 human-collected training data. The proposed method significantly improved the success rate from 40.2\% to 85.7\%, and succeeded the task more than 75\% for other objects.
\end{abstract}


\titlepgskip=-15pt

\maketitle

\section{Introduction}
\label{sec:introduction}
\PARstart{R}{obotic} manipulation has been widely studied to automate daily human tasks~\cite{ref1, ref2, ref3}. Object manipulation involves grasping an object to stabilize it and unrestrained manipulation without grasping the object. The latter unrestrained manipulation is called nonprehensile manipulation, which allows manipulation beyond the range of motion and manipulation of objects that cannot be grasped~\cite{ref4,ref5}. However, nonprehensile manipulation requires a complex dynamic model of the object and the environment, in addition to the relationship between the robot hand and object or tool~\cite{ref6}. Model-based methods that formulate  complex dynamics with mathematical models have been studied~\cite{ref45,ref46}. In~\cite{ref46}, Planning for sliding and moving an object considering friction and for carrying an object with it considering gravity and inertia are very difficult. Fotheremore, these methods assume the use of known object shapes, postures, material, and desired trajectories. Therefore these methods are computationally expensive and difficult to adapt to new objects and environments~\cite{ref7}.

Recently, end-to-end learning that directly learns object manipulation skills from sensor data has been focused. In particular, imitation learning is known to be one of the most data-efficient methods. Imitation learning is supervised learning from motion data that are demonstrated by humans~\cite{ref20,ref21}; therefore, it learns operation skills that require considering the dynamics of the environment and objects.
There are studies of nonprehensile manipulation using imitation learning such as pushing objects~\cite{ref22,ref23}, ball-in-box~\cite{ref24}, and lifting shoes~\cite{ref25}; these approaches learn behaviors from data without explicitly giving physical properties such as friction or shape of the manipulated object and without calculating dynamics. However, the above studies were conducted at a single speed, which is slower than human daily movements. This is because these imitation learning methods predicted the next response value of the robot and gave it as a command value. In general, there is no ideal control system, and there is a delay between the command and response values. Therefore, conventional imitation learning ~\cite{ref22,ref23,ref24,ref25} requires the operation to be very slow to ensure negligible control delay caused during autonomous motion. In particular, fast and dynamic nonprehensile manipulation is significantly affected by this control delay, and task execution is impossible as well.

Bilateral control-based imitation learning has been proposed as a method that can compensate for this control delay~\cite{ref26,ref27}. We indicated that this method can generate variable speed motions that consider  the dynamics between the robot and environment~\cite{ref28,ref29}. Although this method is expected to enable nonprehensile manipulations at high speed and multiple speeds, it is difficult to learn complex dynamics between objects and the environment from sensor data and requires a significant amount of training data. Because  imitation learning requires human demonstration to collect the teacher data, collecting a large amount of teacher data is costly and becomes an obstruction for complex task execution.

In contrast, self-supervised learning has been studied to enhance supervised learning~\cite{ref30,ref31}. Self-supervised learning allows for self-learning without generate and annotate the training data by humans. Self-supervised learning has also been studied in the field of robotics, such as the feature extraction of image recognition~\cite{ref32},  grasping~\cite{ref33,ref34,ref35}, and liquid pouring~\cite{ref36}; however, these studies have focused on geometric changes such as the grasping position and shape of the object, and there has been no study focused on speed, which is a temporal change.
When the nonprehensile motion speed is changed, the dynamics of the object, as well as the robot itself, can change significantly. Although it is very difficult to represent these complex relationships as functions of various speeds, it is expected to be possible to model them with neural networks~(NNs) based on a large amount of data collected by self-supervised learning. 

In this study, we propose self-supervised learning considering motion speed. 
The proposed method improves the success rate by performing the task with a trained NN and using only the success data for self-supervised learning. We performed the task of scooping and transporting a pancake illustrated in Fig.~\ref{fig:fig0}.
The experimental results indicated that the proposed method improved the success rate and the reproducibility of the motion speed as well. Finally, the task success rate was improved from 40.2\% to 85.7\%, and it achieved higher accuracy of time reproduction than humans. The proposed method does not simply linearly expand the trajectory in response to the speed command but learns the dynamics and generates an appropriate motion depending on the motion speed. Furthermore, the proposed method does not explicitly provide the friction and inertia of the pancake, enabling the NN to learn appropriate actions that take them in consideration from the human demonstration. The fact that NN can generate trajectories based on an understanding of the dynamic characteristics of the environment with dynamics is a great academic novelty, and we can expect integration with conventional studies in the control field in the future because changing commands according to speed is similar to designing controllers according to frequency responses. And proposed method is expected to have a significant social impact in terms of the ease of system implementation and the ability to perform difficult tasks. This study is the first approach in imitation learning that makes dynamic behaviors such as high-speed nonprehensile manipulation possible by performing self-supervised learning using bilateral control-based imitation learning and discusses the limitations and future of this method.
The advantages of the proposed method are as follows:

\begin{itemize}
\item {Automatically collects speed-labeled training data and enables self-supervised learning}
\item {Learns multiple speeds and positions nonprehensile manipulation from small amounts of human demonstrations}
\item {Is capable of reproducing speeds that exceed human-given training data}
\end{itemize}

The remainder of this paper is organized as follows. Section~\ref{sec:related work} presents the related work of this study. Section~\ref{sec:control system} explains the robot control system and bilateral control. In Section~\ref{sec:bilateral control-based imitation learning}, we introduce bilateral control-based imitation learning. Section~\ref{sec:Self-supervised Learning Considering Robot Motion Speed} introduces our proposed method of self-supervised learning considering motion speed. Section~\ref{sec:EXPERIMENTandEVALUATION} provides the experimental description, results, and discussion. Finally, in Section~\ref{sec:Conclusion}, we conclude this study and discuss future research topics.

\Figure[t!](topskip=0pt, botskip=0pt, midskip=0pt)[width=85mm]{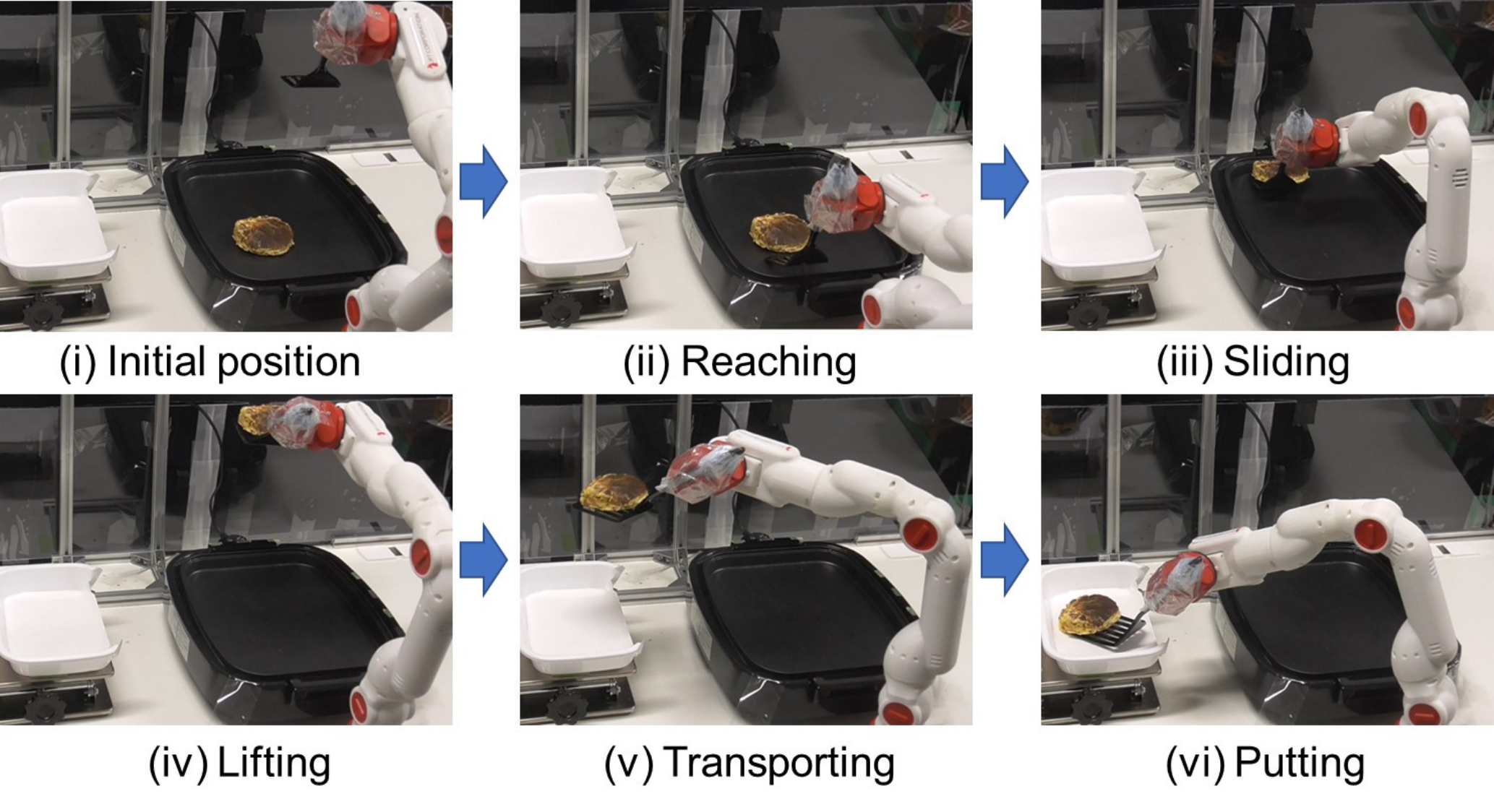}
{Snapshot of scooping and transporting a pancake task \label{fig:fig0}}

\section{RELATED WORK}
\label{sec:related work}
\subsection{Nonprehensile manipulation}
\label{subsec:Nonprehensile manipulation}
Nonprehensile manipulation is performed by not directly grasping the object. Although model-based methods have been studied, there is no general methodology, and nonprehensile manipulation is still a difficult task~\cite{ref46,ref58}. In particular, it is difficult for model-based methods to adapt to change in the state of contact between the object and environment.
Iterative learning control, which performs repetitive motions and corrects  modeling errors, has also been reported~\cite{ref59}.
However, \cite{ref59} assumes  that the desired trajectory is predetermined, which makes it difficult to adapt to changes in the environment.

Recently, model-free methods using machine learning, such as reinforcement learning and imitation learning, have been studied. Reinforcement learning automatically learns behavior based on a reward function through trial and error~\cite{ref8,ref9}. Yuan \textit{et al.} performed the task of pushing and replacing objects~\cite{ref10}, and Finn \textit{et al}. performed the task of scooping up objects with a spatula~\cite{ref11}. However, reinforcement learning requires a huge number of trials to learn a behavior, and it is difficult to set the reward function~\cite{ref12}. In particular, learning nonprehensile manipulation at multiple speeds is inefficient because different reward functions need to be set up and relearned for each speed.
In contrast, imitation learning can be applied with less teacher data because it learns from human motion data. Moreover, it can be easily used by people who are not experts in robotics or machine learning because it does not require reward functions or mathematical models.

\Figure[t!](topskip=0pt, botskip=0pt, midskip=0pt)[width=65 mm]{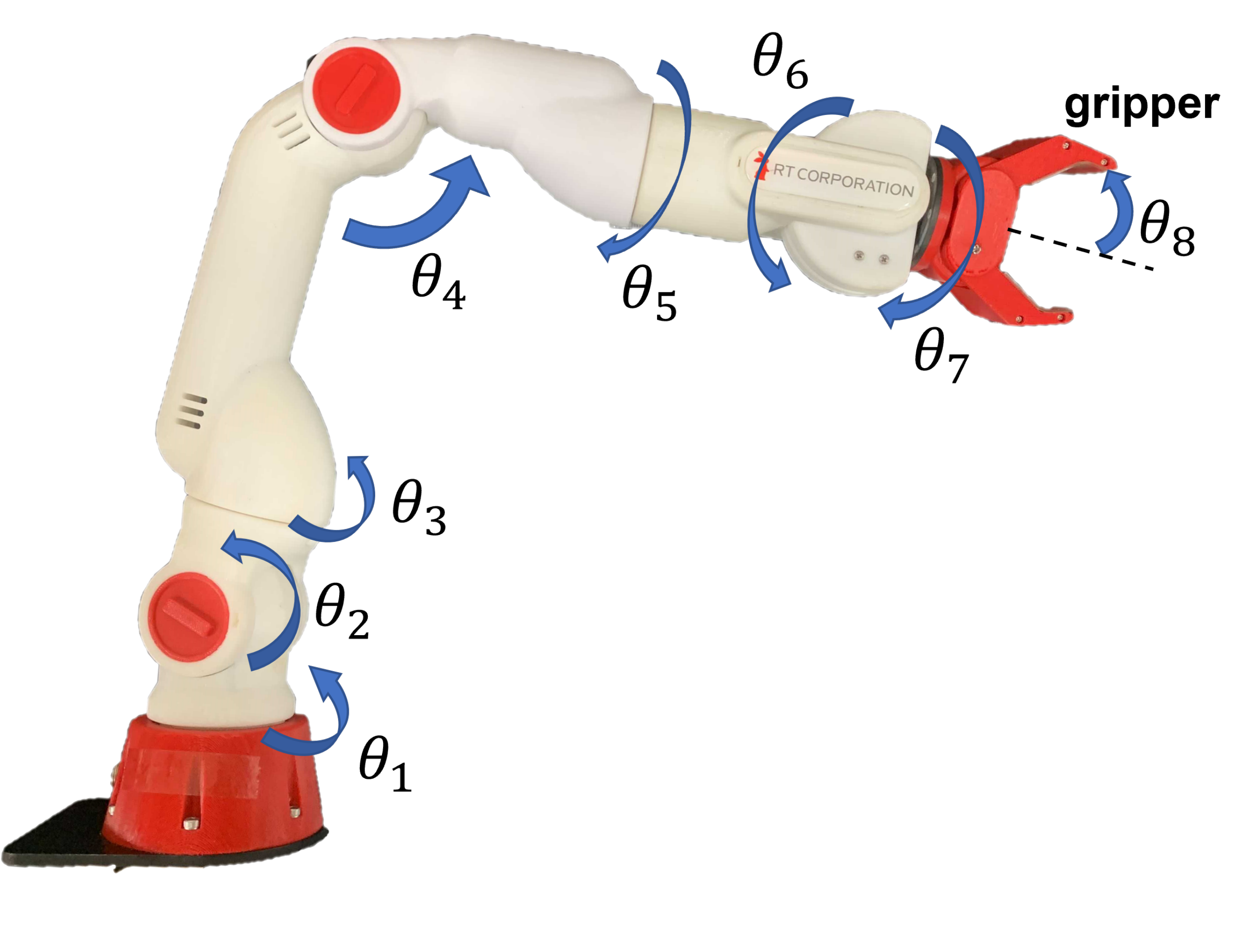}
{Definition~of~robot's~joints (CRANE-X7) \label{fig:fig1}}

\subsection{Imitation learning}
\label{subsec:Imitation learning}
Imitation learning is one of the most efficient methods of end-to-end learning. In the past, several methods using probabilistic models such as hidden Markov models~\cite{ref13} and mixed Gaussian models~\cite{ref14} using NN~\cite{ref15,ref16,ref17} and dynamic movement primitives~\cite{ref18,ref19} have been studied and proved to be effective. However, the movement of these imitation learnings is slower than humans, owing to control delays that occur only during autonomous operations.

Bilateral control-based imitation learning predicts the next command value from the robot's response value and gives that command value to the robot~\cite{ref26,ref27}. This framework of predicting the command value from the response value allows for fast object manipulation considering the control delay. Because this control delay has a significant effect on nonprehensile manipulation, this study employs bilateral control-based imitation learning.

\subsection{Self-supervised learning}
\label{subsec:Self-supervised learning}
Self-supervised learning, mainly spread in the field of image recognition, was used as a pre-training for supervised learning in~\cite{ref47}, and indirectly improved the success rate of the main task by setting sub-tasks in~\cite{ref48}.

Recently, self-supervised learning has also been studied in the field of robotics. Most of the research has been focused on image information, such as object pose estimation~\cite{ref49,ref50}. Huang \textit{et al.} applied self-supervised learning to robot motion data for a task of pouring a specified amount of liquid~\cite{ref36}. In~\cite{ref36}, self-learning based on the actual amount of liquid poured was employed to enable the robot pour a specified amount of liquid, even if it is a new liquid; however, \cite{ref36} deals with the control of one degree of freedom~(DOF) and does not consider the operation time.

Regarding dynamic object manipulation such as nonprehensile manipulation at any speed, it is necessary to consider the interaction with the environment and the dynamics of the manipulated object according to the motion speed. There has been no study on self-supervised learning to learn dynamic behaviors using a time metric.
In this study, we propose a self-supervised learning method using task completion time to achieve nonprehensile manipulation at any speed. The proposed method allows the NN to learn the dynamics of the entire task through the metric of speed. In the field of control engineering, it is possible to understand the dynamics of a linear system by drawing a Bode diagram from input-output data at multiple frequencies. Given that the NN input and output data of nonprehensile operations performed at multiple speeds are similar while using Bode diagrams, it will be possible to learn the dynamics of the entire task. 
Our proposal is different from conventional methods in that it simply changes the parameter of the amount of liquid to motion speed, because it is capable of learning the dynamics of the environment and manipulated objects, and it is possible to generate nonlinear motion in accordance with speed.

\section{CONTROL SYSTEM}
\label{sec:control system}
\subsection{manipulater}
\label{subsec:manipulater}
We utilized two CRANE-X7 manipulators manufactured by RT Corporation, as illustrated in Fig.~\ref{fig:fig1}. This robot has a mechanism of 7-DOF joints and a 1-DOF gripper, and the angle of each joint can be measured. The angles corresponding to each joint $\theta_1, \theta_2, \theta_3, \theta_4, \theta_5, \theta_6, \theta_7$, and gripper $\theta_8$ are defined as per Fig.~\ref{fig:fig1}.

\subsection{controller}
\label{subsec:controller}
The manipulator control system comprised position and force controllers. The position controller comprised a proportional and differential controller, whereas the force controller comprised a proportional controller. The control system is depicted in Fig.~\ref{fig:fig2}. In the figure, $\theta,\dot{\theta}$, and $\tau$ represent the angle, angular velocity, and torque of each joint, respectively, and the superscripts $cmd$, $res$, $ref$, and $dis$ represent the command, response, reference, and disturbance values, respectively. The caret $\hat{\bigcirc}$ indicates the estimated values. The joint angle of each joint was obtained by the optical encoder and angular velocity was calculated by its pseudo-differential. The disturbance torque $\tau^{dis}$ was calculated using a disturbance observer (DOB)~\cite{ref37} and the torque response value $\tau^{res}$ was calculated using a reaction force observer~(RFOB)~\cite{ref38}. Specifically, a force sensor was not utilized in this study. The details of RFOB are presented in Section~\ref{subsec:robot dynamics and sensorless reaction force measurement}.

\subsection{four-channel bilateral control}
\label{subsec:four-channel bilateral control}
Bilateral control is a teleoperation system that utilizes two robots: a leader and a follower. Bilateral control synchronizes the positions of the two robots and presents the reaction force caused by the contact of the follower with the environment to the leader~\cite{ref39}. Using this technique, the human operating the leader can execute tasks as if they were directly controlling the follower. In particular, four-channel bilateral control, which has position and force controllers in both the leader and follower, is the best method for imitation learning using force information~\cite{ref27}. Therefore, four-channel bilateral control was implemented in this study. The block diagram of four-channel bilateral control is showed in Fig.~\ref{fig:fig3}. The angle and torque targets for four-channel bilateral control are defined as follows:
\begin{align}
  \bm{\theta}^{res}_l - \bm{\theta}^{res}_f = & \ \bm{0}, \label{siki:siki1}\\
  \bm{\tau}^{res}_l + \bm{\tau}^{res}_f = & \ \bm{0}. \label{siki:siki2}
\end{align}
where $\bm{\theta}$ is the angle vector, $\bm{\tau}$ is the torque vector, and the subscripts $l$ and $f$ represent the leader and follower, respectively.
Because this study utilized a robot with 8-DOF, including the gripper, the angle response value vector $\bm{\theta}^{\it res}$ is $\bm{\theta}^{\it res}=[{\theta}^{\it res}_{1}, {\theta}^{\it res}_{2}, \cdots, {\theta}^{\it res}_{8}]^T$. The torque response value vector $\bm{\tau}^{\it res}$ is $\bm{\tau}^{\it res}=[{\tau}^{\it res}_{1},{\tau}^{\it res}_{2}, \cdots, {\tau}^{\it res}_{8}]^T$. The torque reference value vectors of the controller that satisfies equations  (\ref{siki:siki1}) and (\ref{siki:siki2}) are represented by the following equations:
\Figure[t!](topskip=0pt, botskip=0pt, midskip=0pt)[width=80mm]{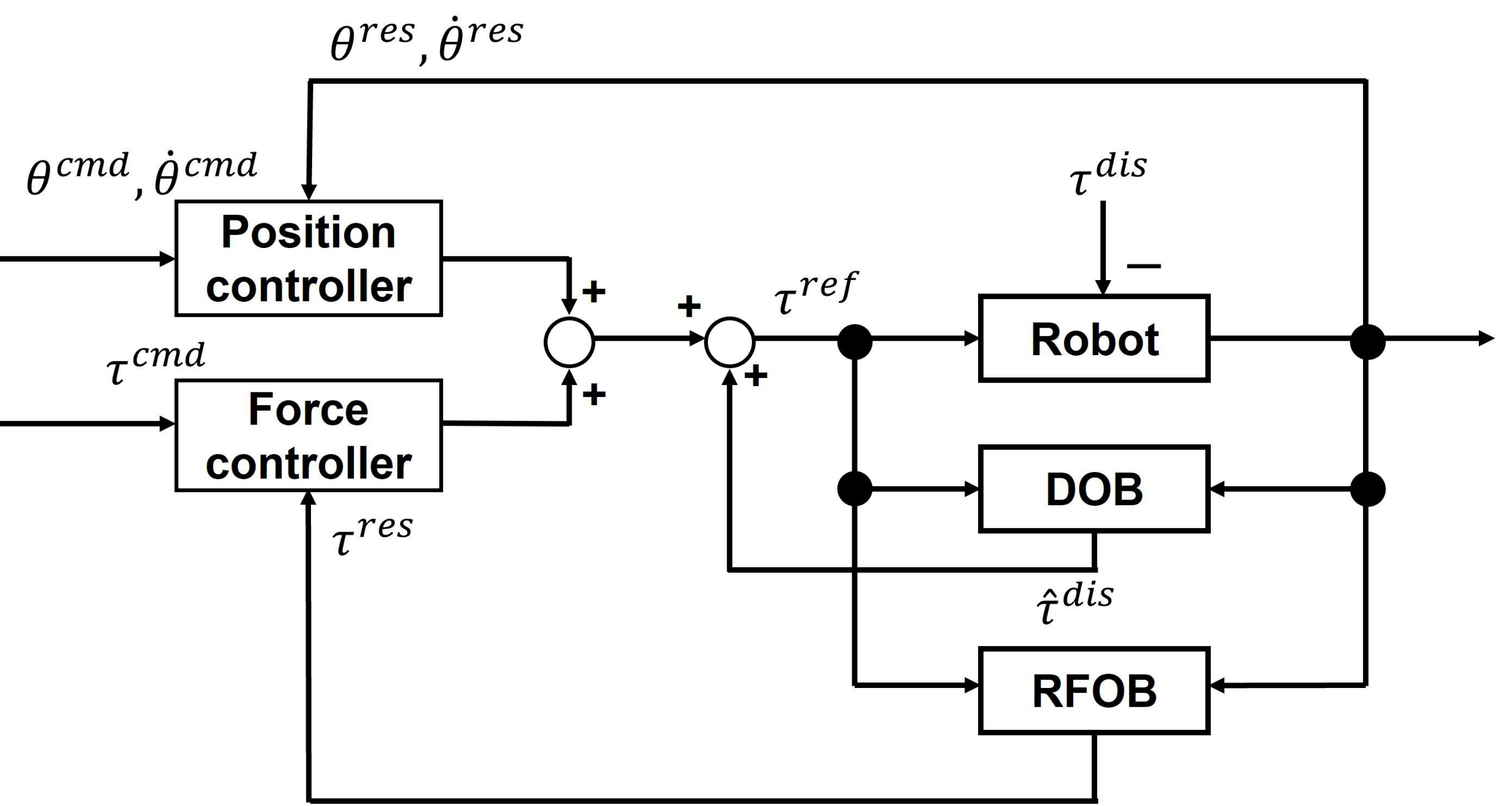}
{Block diagram of controller \label{fig:fig2}}
\begin{align}
  \bm{\tau}^{ref}_l = & \ - \frac{\bm{J}}{2}(\bm{K_p} + \bm{K_d} s)(\bm{\theta}^{res}_l -\bm{\theta}^{res}_f) \nonumber \\ & - \frac{1}{2}\bm{K_f}(\bm{\tau}^{res}_f+\bm{\tau}^{res}_l), \label{siki:siki3}\\
  \bm{\tau}^{ref}_f = & \ \frac{\bm{J}}{2}(\bm{K_p} + \bm{K_d}s)(\bm{\theta}^{res}_l -\bm{\theta^}{res}_f) \nonumber \\ & - \frac{1}{2}\bm{K_f}(\bm{\tau}^{res}_f+\bm{\tau}^{res}_l). \label{siki:siki4}
\end{align}
where $\bm{K_p}$ and $\bm{K_d}$ represent the diagonal gain matrix of proportional and differential control of the position, $\bm{K_f}$ represents the diagonal proportional control gain matrix of force, and $\bm{J}$ is the inertia matrix and $s$ is the Laplace operator. 
These matrix subscripts represent each joint, and the interference of the axes was neglected.
\begin{align}
\bm{K_p} &=\  \rm{diag}[K_{p1},K_{p2}, \cdots, K_{p8}], \label{siki:siki5} \\
\bm{K_d}&=\  \rm{diag}[K_{d1},K_{d2}, \cdots, K_{d8}], \label{siki:siki6} \\
\bm{K_f}&=\  \rm{diag}[K_{f1},K_{f2}, \cdots, K_{f8}], \label{siki:siki7} \\
\bm{J}&=\  \rm{diag}[J_1,J_2,\cdots, J_8].  \label{siki:siki8}
\end{align}
The gain values adopted are listed in Table~\ref{tab:tab1}. The control gains were determined by trial and error to be the maximum value within the range in which the robot does not vibrate. We used the same value for both robots. In this study, $\theta_1$ and $\theta_2$ were controlled with a cutoff frequency of 15.0~rad/s, and the other joints with a cutoff frequency of 20.0~rad/s.
\Figure[t!](topskip=0pt, botskip=0pt, midskip=0pt)[width=80mm]{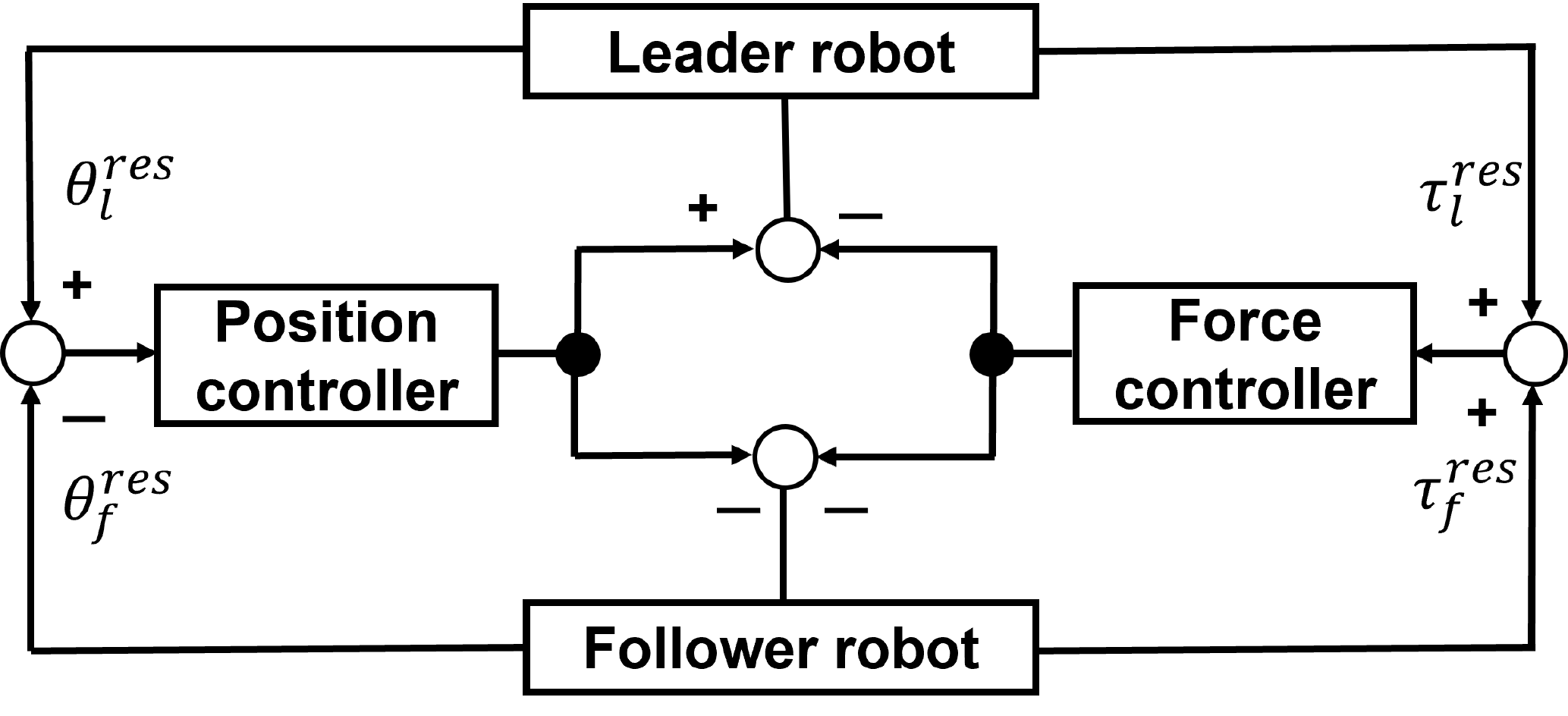}
{Block diagram of four-channel bilateral control \label{fig:fig3}}
\begin{table}[tb]
 \begin{center}
\caption{Gain values for the robot controller}
\begin{tabular}{ccc}
\hline
                     & Parameter                      & Value                    \\ \hline \hline 
$K_{p1}$          & Joint 1's Position feedback gain                    & 256               \\
$K_{p2}$          & Joint 2's Position feedback gain                    & 196               \\
$K_{p3}$          & Joint 3's Position feedback gain                    & 961               \\
$K_{p4}$          & Joint 4's Position feedback gain                    & 225               \\
$K_{p5}$          & Joint 5's Position feedback gain                    & 289               \\
$K_{p6}$          & Joint 6's Position feedback gain                    & 324               \\
$K_{p7}$          & Joint 7's Position feedback gain                    & 144               \\
$K_{p8}$          & Gripper's Position feedback gain                    & 324               \\
$K_{d1}$                   & Joint 1's Velocity feedback gain                   & 32.0                 \\
$K_{d2}$                   & Joint 2's Velocity feedback gain                   & 28.0                 \\
$K_{d3}$                   & Joint 3's Velocity feedback gain                   & 62.0                 \\
$K_{d4}$                   & Joint 4's Velocity feedback gain                   & 30.0                 \\
$K_{d5}$                   & Joint 5's Velocity feedback gain                   & 34.0                 \\
$K_{d6}$                   & Joint 6's Velocity feedback gain                   & 36.0                 \\
$K_{d7}$                   & Joint 7's Velocity feedback gain                   & 24.0                 \\
$K_{d8}$                   & Gripper's Velocity feedback gain                   & 36.0                 \\
$K_{f1}$                   & Joint 1's Force feedback gain                     & 0.70                  \\
$K_{f2}$                   & Joint 2's Force feedback gain                     & 0.70                  \\
$K_{f3}$                   & Joint 3's Force feedback gain                     & 1.00                  \\
$K_{f4}$                   & Joint 4's Force feedback gain                     & 0.75                  \\
$K_{f5}$                   & Joint 5's Force feedback gain                     & 0.80                  \\
$K_{f6}$                   & Joint 6's Force feedback gain                     & 1.00                  \\
$K_{f7}$                   & Joint 7's Force feedback gain                     & 0.80                  \\
$K_{f8}$                   & Gripper's Force feedback gain                     & 1.00                  \\   \hline
\end{tabular}
\label{tab:tab1}
\end{center}
\end{table}

\begin{table}[tb]
 \begin{center}
\caption{Identified system parameters for the robot controller}
\begin{tabular}{ccc}
\hline
                     & Parameter                      & Value                    \\ \hline \hline 
$J_1$                   & Joint 1's inertia {[}N$\cdot$m{]}                     &     0.012                 \\
$J_2$                   & Joint 2's inertia {[}N$\cdot$m{]}                     &     0.113                 \\
$J_3$                   & Joint 3's inertia {[}N$\cdot$m{]}                     &      0.012                \\
$J_4$                   & Joint 4's inertia {[}N$\cdot$m{]}                     &     0.040                 \\
$J_5$                   & Joint 5's inertia {[}N$\cdot$m{]}                     &     0.006                 \\
$J_6$                   & Joint 6's inertia {[}N$\cdot$m{]}                     &      0.007                \\ 
$J_7$                   & Joint 7's inertia {[}N$\cdot$m{]}                     &      0.006                \\
$J_8$                   & Gripper's inertia {[}N$\cdot$m{]}                     &      0.007                \\ 
$D_1$                   & Joint 1's friction compensation cofficient {[}N$\cdot$m{]}                     &     0.050                 \\
$D_2$                   & Joint 3's friction compensation cofficient {[}N$\cdot$m{]}                     &     0.242                 \\
$D_3$                   & Joint 5's friction compensation cofficient {[}N$\cdot$m{]}                     &      0.040                \\
$D_4$                   & Joint 6's friction compensation cofficient {[}N$\cdot$m{]}                     &     0.039                 \\
$D_5$                   & Joint 7's friction compensation cofficient {[}N$\cdot$m{]}                     &     0.050                 \\
$D_6$                   & Gripper's friction compensation cofficient {[}N$\cdot$m{]}                     &     0.021                 \\
$M_{l1}$                   & Leader's joint 2 gravity compensation coefficient 1 {[}N$\cdot$m{]}                     &     2.094                 \\
$M_{l2}$                   & Leader's joint 2 gravity compensation coefficient 2 {[}N$\cdot$m{]}                     &     1.151                 \\
$M_{l3}$                   & Leader's joint 4 gravity compensation coefficient 1 {[}N$\cdot$m{]}                     &     1.183                 \\
$M_{f1}$                   & Follower's joint 2 gravity compensation coefficient 1 {[}N$\cdot$m{]}                     &     2.294                 \\
$M_{f2}$                   & Follower's joint 2 gravity compensation coefficient 2 {[}N$\cdot$m{]}                     &     1.451                 \\
$M_{f3}$                   & Follower's joint 4 gravity compensation coefficient 1 {[}N$\cdot$m{]}                     &     1.483                 \\ \hline
\end{tabular}
\label{tab:tab2}
\end{center}
\end{table}

\subsection{robot dynamics and sensorless reaction force measurement}
\label{subsec:robot dynamics and sensorless reaction force measurement}
In this section, we describe the dynamics of the robot and torque sensorless measurement of reaction force using RFOB.
In this study, the dynamics of the robot is represented as follows:
\begin{align}
J_1\ddot{\theta_1}^{res}  = & \ \tau^{ref}_1 - \tau^{res}_1 - D_1\dot{\theta}_1^{res}, \label{siki:siki9} \\ 
J_2\ddot{\theta_2}^{res}  = & \ \tau^{ref}_2 - \tau^{res}_2  \nonumber \\ & - M_1\sin(\theta_2^{res})  + M_2\sin(\theta_2^{res} + \theta_4^{res}) ,\label{siki:siki10} \\
J_3\ddot{\theta_3}^{res}  = & \ \tau^{ref}_3 - \tau^{res}_3 - D_2\dot{\theta}_3^{res}, \label{siki:siki11} \\
J_4\ddot{\theta_4}^{res}  = & \ \tau^{ref}_4 - \tau^{res}_4 + M_3\sin(\theta_2^{res} + \theta_4^{res}), \label{siki:siki12} \\
J_5\ddot{\theta_5}^{res}  = & \ \tau^{ref}_5 - \tau^{res}_5 - D_3\dot{\theta}_5^{res}, \label{siki:siki13} \\
J_6\ddot{\theta_6}^{res}  = & \ \tau^{ref}_6 - \tau^{res}_6 - D_4\dot{\theta}_6^{res}, \label{siki:siki14} \\
J_7\ddot{\theta_7}^{res}  = & \ \tau^{ref}_7 - \tau^{res}_7 - D_5\dot{\theta}_7^{res}, \label{siki:siki15} \\
J_8\ddot{\theta_8}^{res}  = & \ \tau^{ref}_8 - \tau^{res}_8 - D_6\dot{\theta}_8^{res}.\label{siki:siki16}
\end{align}
where $D$ and $M$ are the friction and gravity compensation coefficients, respectively, and the respective subscripts are adopted to identify each coefficient. The off-diagonal terms in the inertia matrix were considered as negligible. The subscripts of the other parameters correspond to the numbers of the joints. Based on (\ref{siki:siki9})--(\ref{siki:siki16}), the physical parameters were identified. Using maximum length null sequence signal input, measured angular velocity and angular acceleration during free motion of each joint. Then, assuming that $\bm{\tau}^{res}=\bm{0}$, each parameter was determined by the least squares method. The output of DOB, calculated from the torque reference and response values of acceleration, is represented as follows:
\begin{gather}
\hat{\bm{\tau}}^{dis} = \bm{\tau}^{ref} - \bm{\tau}^{res} - \bm{J}\ddot{\bm{\theta}}^{res}, \label{siki:siki17}
\end{gather}
Using the identified parameters and (\ref{siki:siki9})--(\ref{siki:siki16}), the reaction force was calculated from RFOB without force sensor as expressed in the following equation.
\begin{align}
\tau^{res}_1 = & \ \tau^{dis}_1 - D_1\dot{\theta}_1^{res}, \label{siki:siki18} \\
\tau^{res}_2 = & \ \tau^{dis}_2 - M_1\sin(\theta_2^{res}) + M_2\sin(\theta_2^{res} + \theta_4^{res}), \label{siki:siki19} \\
\tau^{res}_3 = & \ \tau^{dis}_3 - D_2\dot{\theta}_3^{res}, \label{siki:siki20} \\
\tau^{res}_4 = & \ \tau^{dis}_4 + M_3\sin(\theta_2^{res} + \theta_4^{res}), \label{siki:siki21} \\
\tau^{res}_5 = & \ \tau^{dis}_5 - D_3\dot{\theta}_5^{res}, \label{siki:siki22} \\
\tau^{res}_6 = & \ \tau^{dis}_6 - D_4\dot{\theta}_6^{res}, \label{siki:siki23} \\
\tau^{res}_7 = & \ \tau^{dis}_7 - D_5\dot{\theta}_7^{res}, \label{siki:siki24} \\
\tau^{res}_8 = & \ \tau^{dis}_8 - D_6\dot{\theta}_8^{res}. \label{siki:siki25}
\end{align}

The physical parameters of the robot are listed in Table~\ref{tab:tab2}.
In this study, the turner was fixed on the follower, thus the gravity compensation value of the follower was increased.

 \Figure[t!](topskip=0pt, botskip=0pt, midskip=0pt)[width=70mm]{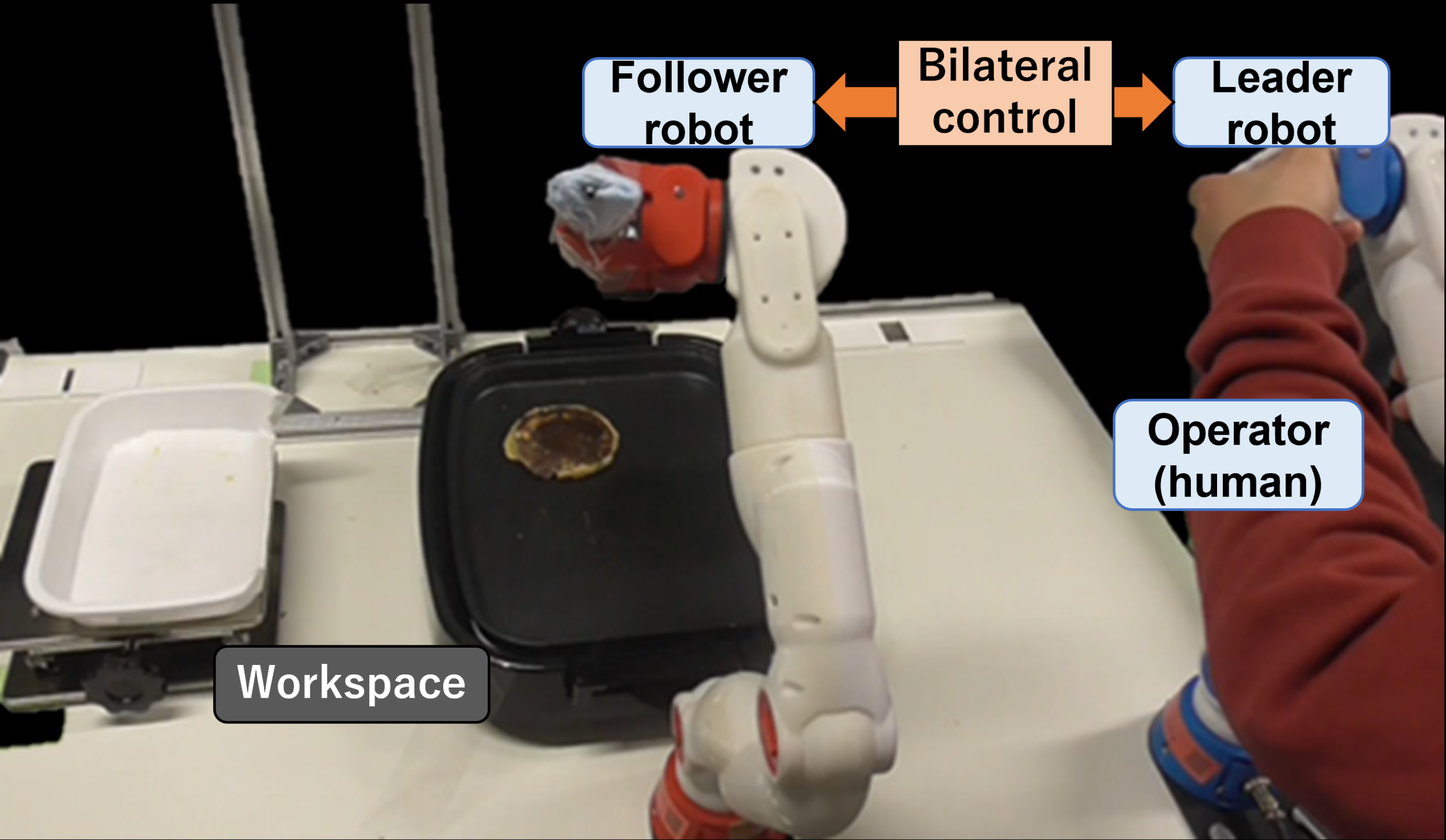}
{Collection of training data using bilateral control \label{fig:fig6}}

\section{Bilateral control-based imitation learning}
\label{sec:bilateral control-based imitation learning}
Bilateral control-based imitation learning can imitate the subtle force of humans and  manipulate objects faster than other imitation learning methods. In this section, the flow of bilateral control-based imitation learning is divided into three phases.
\subsection{Collecting training data}
\label{subsec:collecting training data}
 In bilateral control-based imitation learning, two robots were utilized only when collecting training data. 
We experimented with nonprehensile manipulation by scooping up a pancake on a hot plate, and thereafter transporting and placing it on an adjacent tray at multiple speeds. Therefore, operator demonstrated nonprehensile manipulation at multiple speeds using bilateral control. The operator executed the task by controlling the leader and remotely controlling the follower that was in the workspace, as illustrated in Fig.~\ref{fig:fig6}. 
Because the robot in Fig.~\ref{fig:fig1} had a redundant DOF, $\theta_3$ was fixed with position control, and the robot was controlled at a frequency of 500~Hz, which was the fastest control cycle for this robot. 

\subsection{Training the NN model}
\label{subsec:training the NN model}
In bilateral control-based imitation learning, using NN as learning model. The basic structure of the NN for bilateral control-based imitation learning is displayed in Fig.~\ref{fig:fig4}. 

In bilateral control-based imitation learning, the model is constructed with the response value of the follower as the input and that of the leader as the output. This configuration makes it possible to reproduce the same system in autonomous operation as during training data collection, but without the human and leader robot. The difference between bilateral control during collecting training data and executing a task is illustrated in Fig.~\ref{fig:fig5}. Recurrent Neural Network ~(RNN)~\cite{ref22,ref26,ref27,ref28,ref29,ref36,ref17} and Convolutional Neural Network~(CNN)~\cite{ref22,ref23,ref25,ref15,ref16} are mainly used in the research of robot motion generation using NNs. RNN was suitable for learning time series information, while CNN was suitable for learning image information. In this study, we used RNN because we did not use image information for training, and among RNN, we used Long Short-Term Memory~(LSTM), which can handle long time-series data.

\Figure[t!](topskip=0pt, botskip=0pt, midskip=0pt)[width=80mm]{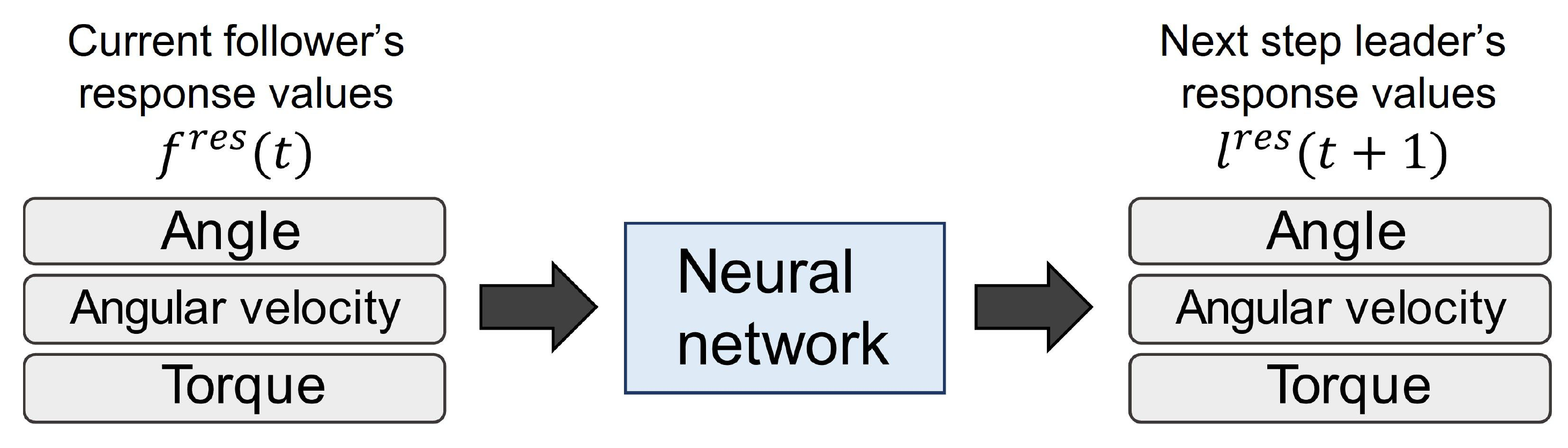}
{Basic structure of learning model in bilateral control-based imitation learning \label{fig:fig4}}

\subsection{Executing a task}
\label{subsec:executing a task}
During autonomous operation, the trained NN replaced the human and the leader robot.
In this study, the inference of NN was done every 20~ms, and the control cycle of the robot was 2~ms, which was the same as that when collecting the training data.

\section{Self-supervised Learning Considering Speed}
\label{sec:Self-supervised Learning Considering Robot Motion Speed}
\begin{figure*}[tb]
    \centering
        \includegraphics[width=140mm]{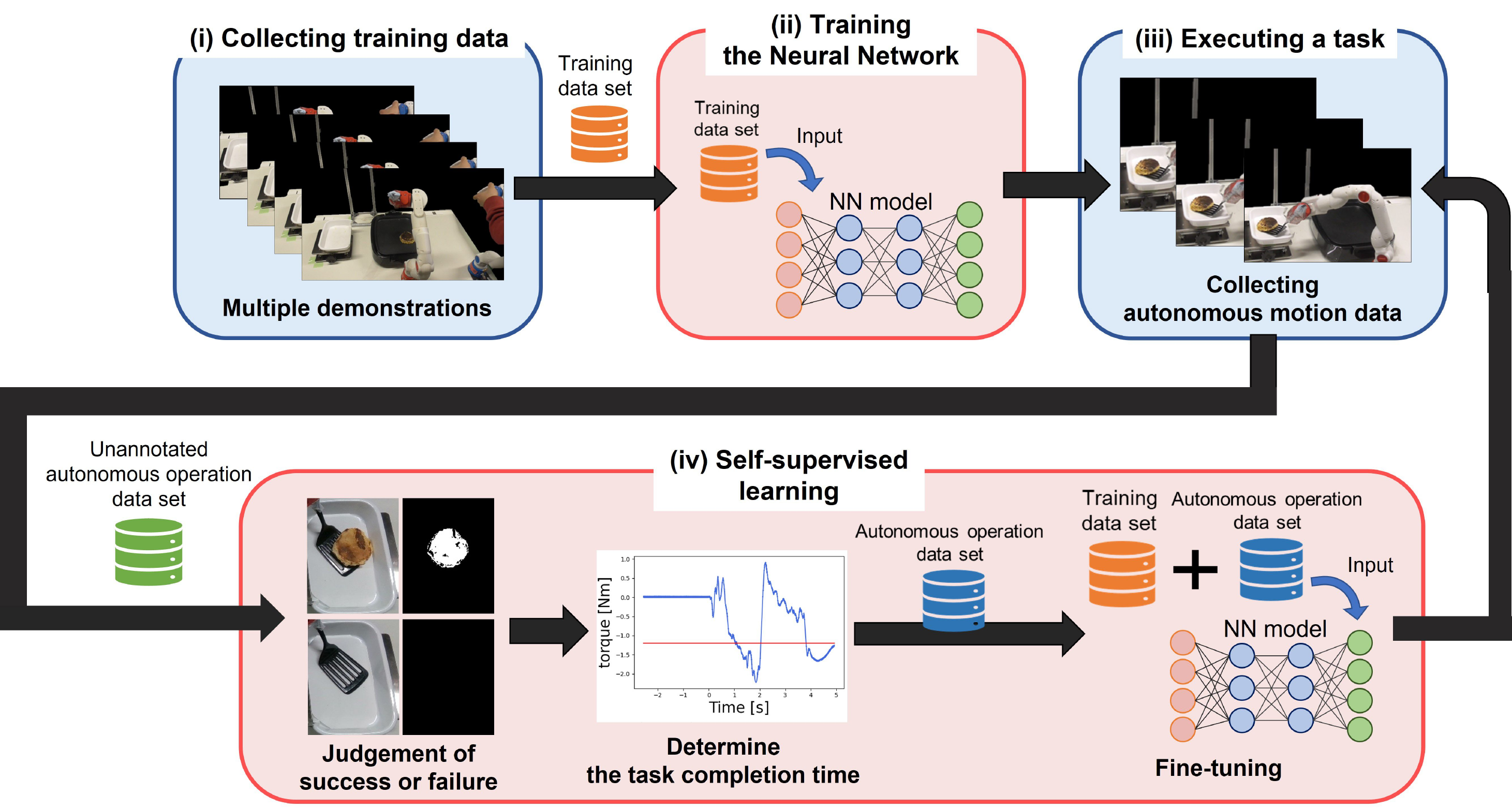}
        \caption{Flowchart of the proposed method.~(i)--(iii) are bilateral control-based imitation learning, and (iv)~is the proposed method. \newline (iv) includes success/failure determination, teacher label generation, and fine-tuning.}
        \label{fig:fig7}
\end{figure*}

Our self-supervised learning fine-tunes the NN using data of autonomous operations conducted in Section~\ref{subsec:executing a task}. Therefore, this method is regarded as the fourth stage following those described in Section~\ref{sec:bilateral control-based imitation learning}: collecting training data, training the NN model, and executing a task. The process from bilateral control-based imitation learning to self-supervised learning of the proposal is illustrated in Fig.~\ref{fig:fig7}. When a task is executed using an NN that has not been overtrained, the behavior in each trial is different as though it is similar. In our method, trying the task multiple times and using only the successful motions within the variance for learning makes it possible to learn only the good behaviors from the minute differences. Repeating this process at multiple speeds will enable progressive learning of the dynamics of the entire task and improve the success rate.

To reuse the motion data during autonomous operation, it is necessary to determine the success or failure of the task, as well as annotating the generated behavior.
In the following, we explain the automated annotation method, the method of determining success or failure, and the method of creating a dataset for relearning.

\subsection{Determine the task completion time}
\label{subsec:Calculation of task completion time}
In this study, the task was to scoop up a pancake on a hot plate and place it on an adjacent tray. Therefore, the time when the torque response value of $\theta_2$, triggered by the movement to place the pancake on the tray was observed above a threshold, was defined as the task completion time.
The angle response value of $\theta_1$ and torque response value of $\theta_2$ during task execution are illustrated in Fig.~\ref{fig:fig12}. The red line in Fig.~\ref{fig:fig12}~(b) is 1.2~N$\cdot$m, which was set as the threshold in this study, and $t_f$ is the task completion time. The sliding and putting phases are distinguished using a threshold based on the angular response value of $\theta_1$ in Fig.~\ref{fig:fig12}~(a). Based on the accurately labeled speed, the NN corrects the error of the task completion time.

\Figure[t!](topskip=0pt, botskip=0pt, midskip=0pt)[width=80mm]{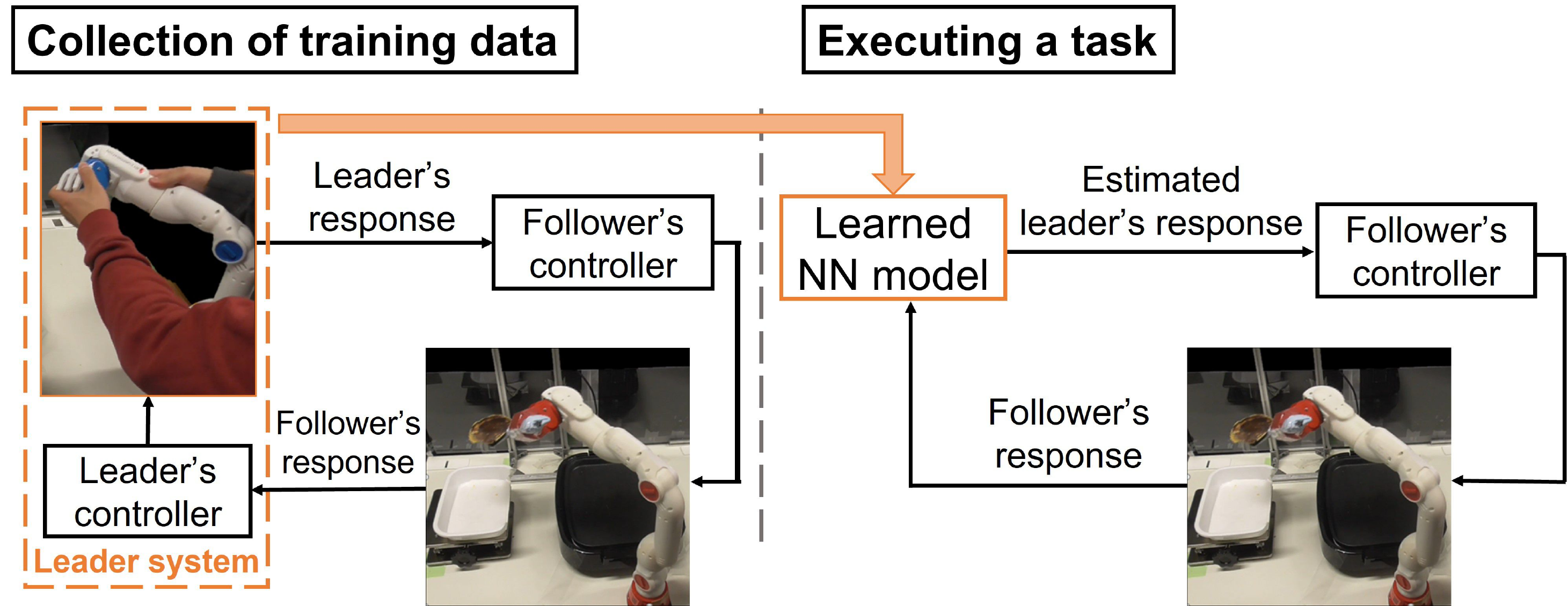}
{Difference in bilateral control when collecting training data and  executing a task \label{fig:fig5}}

\Figure[t!](topskip=0pt, botskip=0pt, midskip=0pt)[width=80mm]{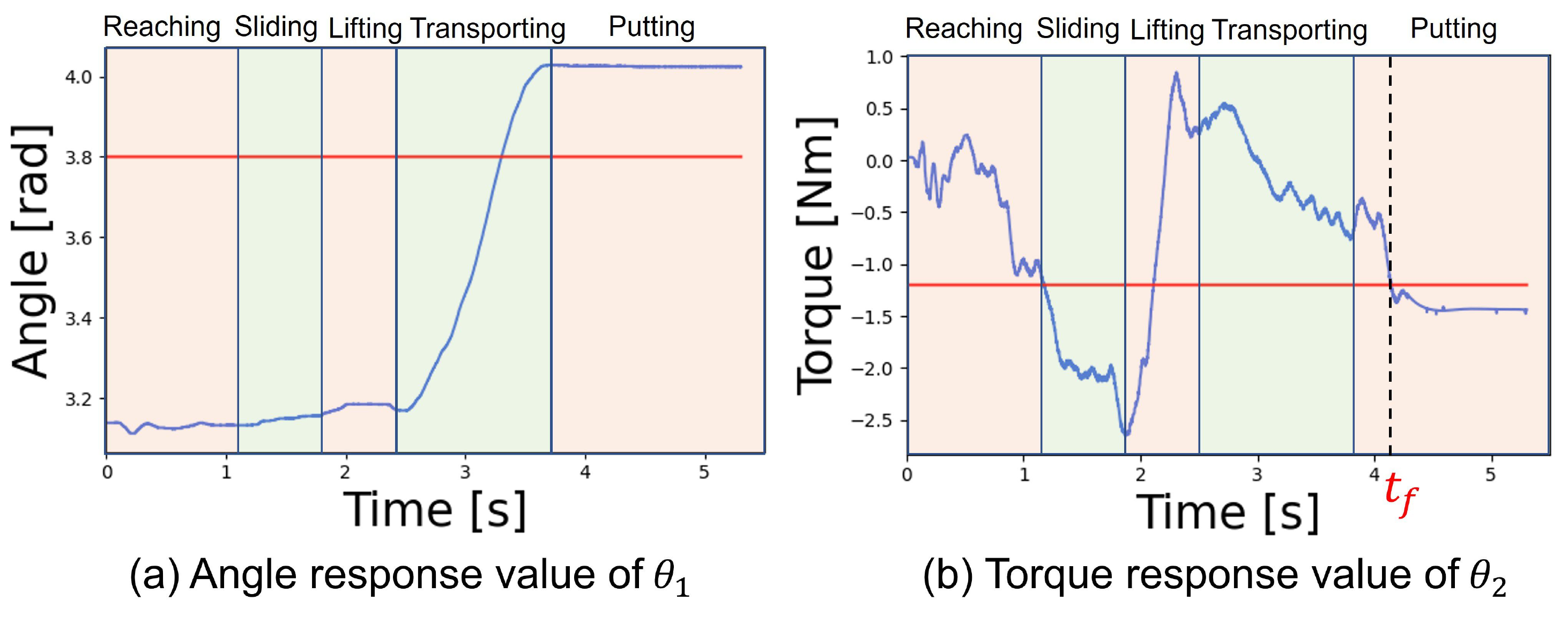}
{Definition of Task Completion Time \label{fig:fig12}}

\subsection{Judging the success or failure of a task}
\label{subsec:Judging the success or failure of a task}
The image on the tray was acquired at the end of the task, and after binarization using HSV color space, the area of the yellow object was calculated to determine its success or failure. An example of the binarization is illustrated in Fig.~\ref{fig:fig8}~(b). As the threshold for binarization using HSV color space, pixels in the range of 15~<~h~<~35.5, 100~<~s, and v~<~180 were set as white. Five hundred  pixels or more of the binarized image were considered successful.

\subsection{Generate position command}
\label{subsec:Generate position command}
To demonstrate that the proposed method can be employed with self-supervised learning focused on spatial information, we considered the initial position of the pancake.
The image of the workspace at the start was obtained, image was binarized by color as in Section~\ref{subsec:Judging the success or failure of a task}, coordinates of the center on the pixel were calculated, and these were adopted as the position coordinates (x,y) of the pancake. The calculated position coordinates were adopted as position command values. Fig.~\ref{fig:fig8} illustrates the coordinates were calculated.

\subsection{Creating a data set for retraining}
\label{subsec:Creating a data set for retraining}
In self-supervised learning, there is no difference between a method that repeatedly collects and trains a small amount of data and a method that collects and trains a large amount of data at once~\cite{ref36}. Therefore, we assume that only a significant amount of data was collected and trained at a time. In short, we performed several actions with the same model and added the obtained action data to the original training dataset to create a new dataset.

\section{EXPERIMENT \& EVALUATION}
\label{sec:EXPERIMENTandEVALUATION}
We examined the task of scooping and transporting a pancake, and this task has different difficulties in each phase. In the reaching and sliding phases, the robot needed to learn sufficient spatial information, because the robot needed to reach an appropriate position and then slide it to the edge of the hot plate. In the lifting phase, the turner needed to be pressed against the edge of the hot plate with a proper force and lifted at the right angle and angular velocity. If the pancake is lifted more than necessary, the pancake will be thrown away; however, if it is not lifted sufficiently, the pancake will slip off the turner. In the transporting phase, it was necessary to move the pancake in an appropriate trajectory that considers inertia and friction. In particular, at high speeds, if the turner is not properly tilted, the pancake will fall in the opposite direction.

We inspected whether repeating the loop of self-supervised learning and executing a task twice, as illustrated in Fig.~\ref{fig:fig7}, would improve the success rate for failed positions and speeds, as well as the reproducibility of task completion times. Moreover, as an additional experiment, we conducted tests on untrained objects to investigate whether generalization performance could be obtained.

\subsection{Design \& Setup}
\label{subsec:DesignandSetup}
The experimental environment is illustrated in Fig.~\ref{fig:fig9}~(a). The camera was an Intel RealSense~D415. The turner was fixed to the robot hand. In the human-collected training data condition, the pancakes were located in four locations: lower left, upper left, lower right, and upper right, as illustrated in Fig.~\ref{fig:fig9}~(b). In addition, the task completion times were 4, 8, and 12~s, and the mass of the pancake was 30~g and 90~g. The pancakes were uniform in size, illustrated in Fig.~\ref{fig:fig9}~(c). In all these combinations, we collected the teacher data twice each. Thus, the number of teacher data was 48  (4~[position] × 3~[completion time] × 2~[mass] × 2~[trial]). Among these, 24~data  were used as training data, and the remaining 24~data were used as validation data. The collected teacher data was downsampled to 50~Hz data with \cite{ref53}, and the number of training data was made 10 times larger.

Autoregressive learning is less affected by covariate shifts and can be more efficient; therefore, the S2SM model of \cite{ref52} was implemented.
The NN utilized in this study is illustrated in Fig.~\ref{fig:fig10}. It comprises a LSTM~8~layers and follows a fully connected layer. The number of layers of NNs was set to a same number with the reference~\cite{ref29}, which achieved variable speed motion generation with contact with the environment. However, the number of units was set to 200 to enable learning of more complex expressions and to avoid a significant increase in computation time, since this study learned a 7-DOF joint, whereas the 3-DOF joint in~\cite{ref29} was learned. Max-min normalization was applied to all 24 dimensions of the input information, and mean squared error was adopted as the loss function. In addition, we utilized Adam~\cite{ref57} as the optimizer, and mini-batch training was applied to a batch of 100~data.

\Figure[t!](topskip=0pt, botskip=0pt, midskip=0pt)[width=80mm]{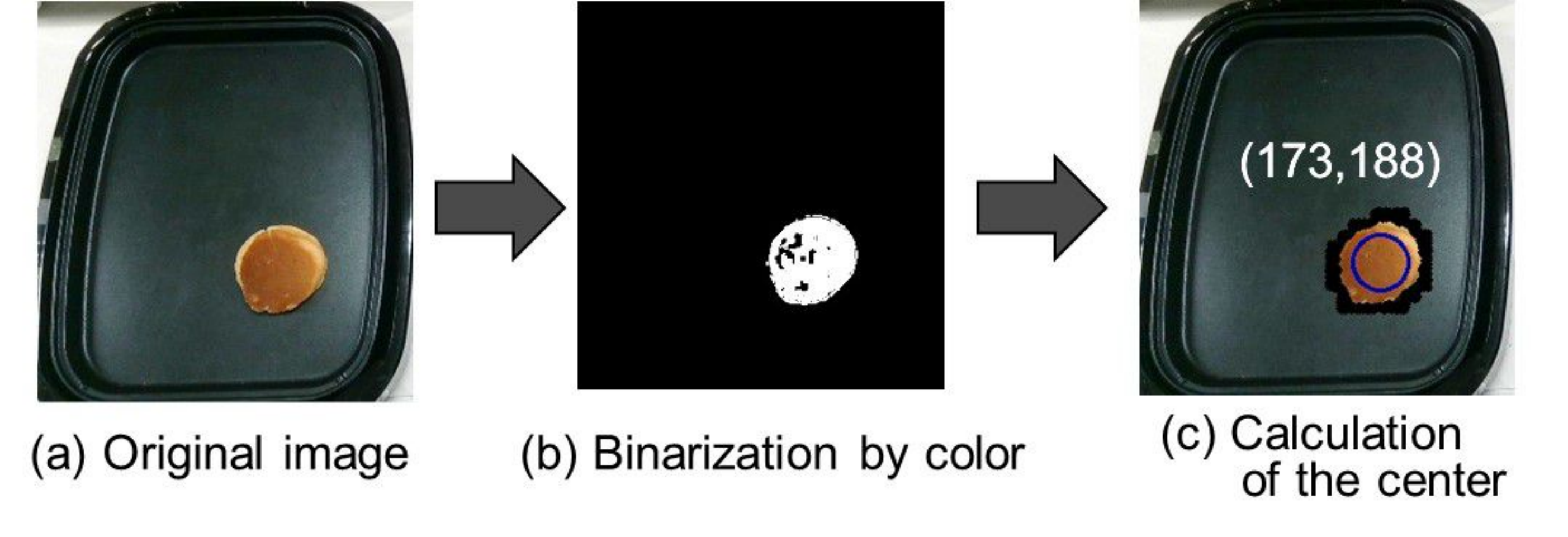}
{Determination of the center coordinates of the pancake \label{fig:fig8}}

\Figure[t!](topskip=0pt, botskip=0pt, midskip=0pt)[width=80mm]{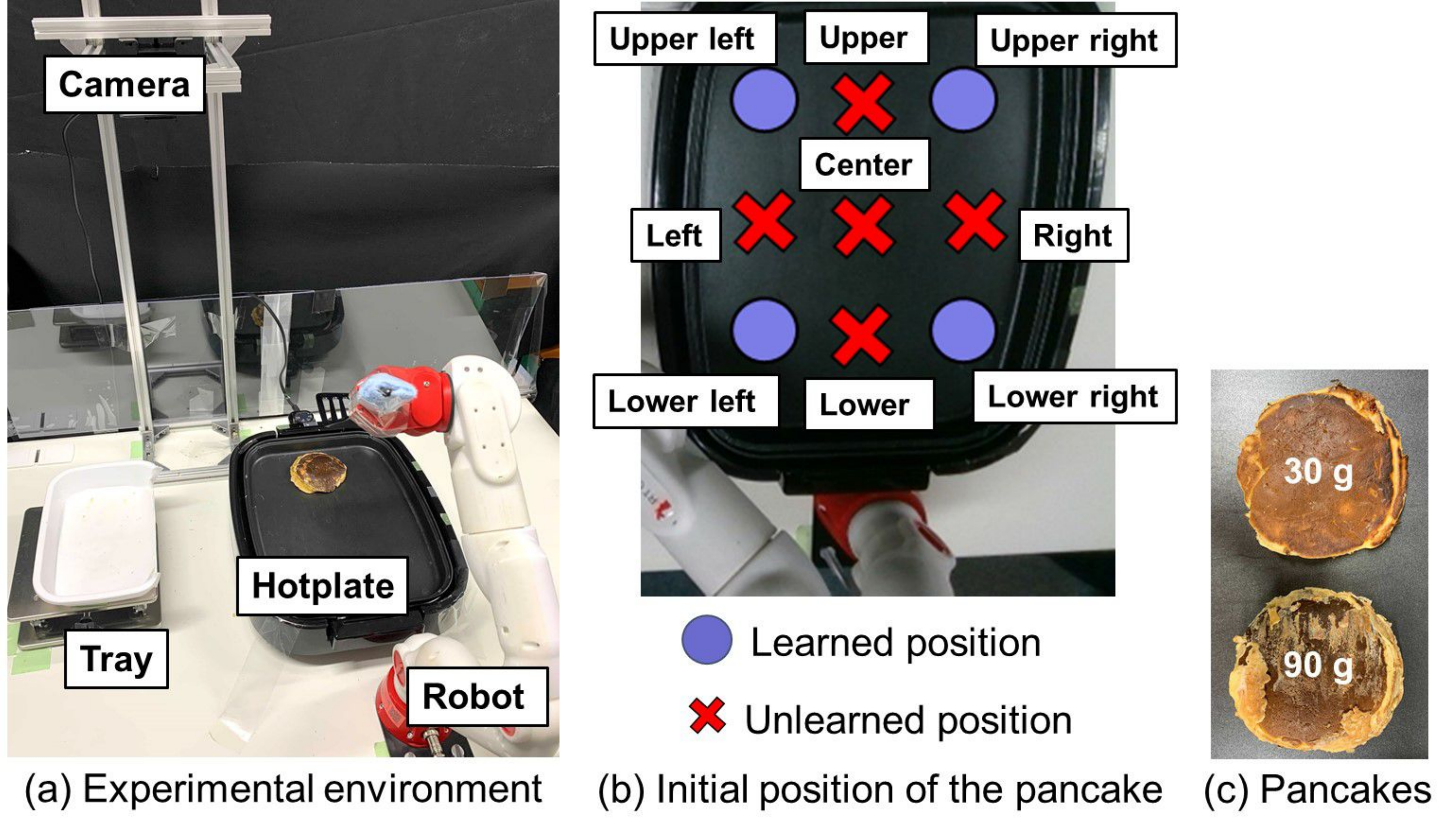}
{Experiment setup \label{fig:fig9}}


\subsection{PRELIMINARY EXPERIMENT}
\label{subsec:PRELIMINARY EXPERIMENT}
\subsubsection{Description}
\label{subsubsec:Description1}
As a preliminary experiment, we performed the task with the original model learned using 24~training data. The test was performed thrice for each combination of all 9~positions illustrated in Fig.~\ref{fig:fig9}~(b) and 11~speeds (3, 4, 5, 6, 7, 8, 9, 10, 11, 12, 13~s). Therefore, the total number of tests was 594 (9~[position] × 11~[completion time] × 2~[mass] × 3~[trial]).
The number of training times of the NN was determined regarding  the validation loss, and 8000~times was adopted. This model is referred to as the original model. The loss graph for the original model is illustrated in  Fig.~\ref{fig:fig15}~(a)

\subsubsection{Result}
\label{subsubsec:Result1}
The task success rates for the original model are presented in Table~\ref{tab:tab3}. Because no significant difference was observed in the results depending on the mass of the pancake, a summary table is presented here. The individual results are provided in the appendix. The success rate of the original model was 40.2\% and the task success rates at untrained positions were remarkably low. We determined that successes and failures depend on the task completion time command even if the position command was the same. 
Most of the failures of the original model were that it operated in a different position from what was commanded. Furthermore, the pancake slipped off when it started to be transported at high speed, and was thrown away during the lifting phase.

Fig.~\ref{fig:fig11} illustrates the reproducibility of completion time. In this figure, the black line is the ideal line, and only successful trials are plotted. Therefore, the position command of Upper with a task completion time of 4~s is not adopted to calculate the mean and variance. From Fig.~\ref{fig:fig11}~(a), we can observe that the unlearned speeds of 5 to 6~s and 10 to 11~s are far from the commanded values, and the variance is also very large. The task completion time entirely tended to be slower than the command time.


\begin{table*}[]
\centering
\caption{Task success rate of the original model}
\scalebox{0.87}{ 
\begin{tabular}{c||ccccccccc||c}
\hline \hline
\multirow{3}{*}{\begin{tabular}[c]{@{}c@{}}Task completion\\ time command {[}s{]}\end{tabular}} & \multicolumn{9}{c|}{Position command}                                                                                                                                                                                                                                                                                                                   & \multirow{3}{*}{\begin{tabular}[c]{@{}c@{}}Speed\\ total\end{tabular}} \\ \cline{2-10}
                                                                                                & \multicolumn{4}{c|}{Learned}                                                                                                                                         & \multicolumn{5}{c|}{Unlearned}                                                                                                                                                   &                                                                        \\ \cline{2-10}
                                                                                                & \multicolumn{1}{c|}{Lower left}         & \multicolumn{1}{c|}{Upper left}        & \multicolumn{1}{c|}{Lower right}        & \multicolumn{1}{c|}{Upper right}        & \multicolumn{1}{c|}{Center}             & \multicolumn{1}{c|}{Left}              & \multicolumn{1}{c|}{Lower}              & \multicolumn{1}{c|}{Right}              & Upper     &                                                                        \\ \hline
3.00                                                                                            & \multicolumn{1}{c|}{0.0(0/6)}           & \multicolumn{1}{c|}{50.0(3/6)}         & \multicolumn{1}{c|}{0.0(0/6)}           & \multicolumn{1}{c|}{\textbf{100(6/6)}}  & \multicolumn{1}{c|}{0.0(0/6)}           & \multicolumn{1}{c|}{0.0(0/6)}          & \multicolumn{1}{c|}{\textbf{100(6/6)}}  & \multicolumn{1}{c|}{0.0(0/6)}           & 0.0(0/6)  & 27.8(15/54)                                                            \\ \hline
4.00                                                                                            & \multicolumn{1}{c|}{$\ast$16.7(1/6)}          & \multicolumn{1}{c|}{$\ast$\textbf{100(6/6)}} & \multicolumn{1}{c|}{$\ast$50.0(3/6)}          & \multicolumn{1}{c|}{$\ast$66.7(4/6)}          & \multicolumn{1}{c|}{\textbf{100(6/6)}}  & \multicolumn{1}{c|}{\textbf{100(6/6)}} & \multicolumn{1}{c|}{\textbf{100(6/6)}}  & \multicolumn{1}{c|}{0.0(0/6)}           & 0.0(0/6)  & 59.3(32/54)                                                            \\ \hline
5.00                                                                                            & \multicolumn{1}{c|}{50.0(3/6)}          & \multicolumn{1}{c|}{50.0(3/6)}         & \multicolumn{1}{c|}{0.0(0/6)}           & \multicolumn{1}{c|}{\textbf{100(6/6)}}  & \multicolumn{1}{c|}{\textbf{100(6/6)}}  & \multicolumn{1}{c|}{66.7(4/6)}         & \multicolumn{1}{c|}{66.7(4/6)}          & \multicolumn{1}{c|}{16.7(1/6)}          & 0.0(0/6)  & 50.0(27/54)                                                            \\ \hline
6.00                                                                                            & \multicolumn{1}{c|}{0.0(0/6)}           & \multicolumn{1}{c|}{0.0(0/6)}          & \multicolumn{1}{c|}{0.0(0/6)}           & \multicolumn{1}{c|}{\textbf{100(6/6)}}  & \multicolumn{1}{c|}{0.0(0/6)}           & \multicolumn{1}{c|}{50.0(3/6)}         & \multicolumn{1}{c|}{66.7(4/6)}          & \multicolumn{1}{c|}{16.7(1/6)}          & 0.0(0/6)  & 25.9(14/54)                                                            \\ \hline
7.00                                                                                            & \multicolumn{1}{c|}{66.7(4/6)}          & \multicolumn{1}{c|}{0.0(0/6)}          & \multicolumn{1}{c|}{0.0(0/6)}           & \multicolumn{1}{c|}{\textbf{100(6/6)}}  & \multicolumn{1}{c|}{0.0(0/6)}           & \multicolumn{1}{c|}{0.0(0/6)}          & \multicolumn{1}{c|}{0.0(0/6)}           & \multicolumn{1}{c|}{66.7(4/6)}          & 0.0(0/6)  & 25.9(14/54)                                                            \\ \hline
8.00                                                                                            & \multicolumn{1}{c|}{$\ast$\textbf{83.3(5/6)}} & \multicolumn{1}{c|}{$\ast$0.0(0/6)}          & \multicolumn{1}{c|}{$\ast$\textbf{83.3(5/6)}} & \multicolumn{1}{c|}{$\ast$\textbf{100(6/6)}}  & \multicolumn{1}{c|}{0.0(0/6)}           & \multicolumn{1}{c|}{0.0(0/6)}          & \multicolumn{1}{c|}{66.7(4/6)}          & \multicolumn{1}{c|}{16.7(1/6)}          & 0.0(0/6)  & 38.9(21/54)                                                            \\ \hline
9.00                                                                                            & \multicolumn{1}{c|}{50.0(3/6)}          & \multicolumn{1}{c|}{0.0(0/6)}          & \multicolumn{1}{c|}{66.7(4/6)}          & \multicolumn{1}{c|}{\textbf{100(6/6)}}  & \multicolumn{1}{c|}{0.0(0/6)}           & \multicolumn{1}{c|}{50.0(3/6)}         & \multicolumn{1}{c|}{0.0(0/6)}           & \multicolumn{1}{c|}{\textbf{83.3(5/6)}} & 0.0(0/6)  & 38.9(21/54)                                                            \\ \hline
10.0                                                                                            & \multicolumn{1}{c|}{16.7(1/6)}          & \multicolumn{1}{c|}{0.0(0/6)}          & \multicolumn{1}{c|}{50.0(3/6)}          & \multicolumn{1}{c|}{\textbf{100(6/6)}}  & \multicolumn{1}{c|}{\textbf{83.3(5/6)}} & \multicolumn{1}{c|}{16.7(1/6)}         & \multicolumn{1}{c|}{66.7(4/6)}          & \multicolumn{1}{c|}{\textbf{100(6/6)}}  & 0.0(0/6)  & 48.1(26/54)                                                            \\ \hline
11.0                                                                                            & \multicolumn{1}{c|}{0.0(0/6)}           & \multicolumn{1}{c|}{\textbf{100(6/6)}} & \multicolumn{1}{c|}{\textbf{83.3(5/6)}} & \multicolumn{1}{c|}{\textbf{83.3(5/6)}} & \multicolumn{1}{c|}{\textbf{83.3(5/6)}} & \multicolumn{1}{c|}{0.0(0/6)}          & \multicolumn{1}{c|}{\textbf{83.3(5/6)}} & \multicolumn{1}{c|}{50.0(3/6)}          & 33.3(2/6) & 57.4(31/54)                                                            \\ \hline
12.0                                                                                            & \multicolumn{1}{c|}{$\ast$0.0(0/6)}           & \multicolumn{1}{c|}{$\ast$\textbf{100(6/6)}} & \multicolumn{1}{c|}{$\ast$0.0(0/6)}           & \multicolumn{1}{c|}{$\ast$66.7(4/6)}          & \multicolumn{1}{c|}{\textbf{83.3(5/6)}} & \multicolumn{1}{c|}{0.0(0/6)}          & \multicolumn{1}{c|}{66.7(4/6)}          & \multicolumn{1}{c|}{66.7(4/6)}          & 0.0(0/6)  & 42.6(23/54)                                                            \\ \hline
13.0                                                                                            & \multicolumn{1}{c|}{0.0(0/6)}           & \multicolumn{1}{c|}{\textbf{100(6/6)}} & \multicolumn{1}{c|}{0.0(0/6)}           & \multicolumn{1}{c|}{\textbf{83.3(5/6)}} & \multicolumn{1}{c|}{0.0(0/6)}           & \multicolumn{1}{c|}{0.0(0/6)}          & \multicolumn{1}{c|}{0.0(0/6)}           & \multicolumn{1}{c|}{66.7(4/6)}          & 0.0(0/6)  & 27.8(15/54)                                                            \\ \hline \hline
\begin{tabular}[c]{@{}c@{}}Position\\ total\end{tabular}                                        & \multicolumn{1}{c|}{25.8(17/66)}        & \multicolumn{1}{c|}{45.5(30/66)}       & \multicolumn{1}{c|}{30.3(20/66)}        & \multicolumn{1}{c|}{\textbf{90.9(60/66)}}        & \multicolumn{1}{c|}{40.9(27/66)}        & \multicolumn{1}{c|}{25.8(17/66)}       & \multicolumn{1}{c|}{56.1(37/66)}        & \multicolumn{1}{c|}{43.9(29/66)}        & 3.0(2/66) & 40.2(239/594)                                                          \\ \hline \hline
\end{tabular}
}
\raggedright
$\ast$:Learned position and speed commands
\label{tab:tab3}
\end{table*}

\subsection{Self-supervised learning(1st time)}
\label{subsec:Self-supervised learning(1st time)}
\subsubsection{Description}
\label{subsubsec:Description2}
To evaluate the effectiveness of self-supervised learning, we conducted an experiment using the autonomous action data from Section~\ref{subsec:PRELIMINARY EXPERIMENT}. For all patterns that were successful at least once during the trials in Section~\ref{subsec:PRELIMINARY EXPERIMENT}, we collected additional data so that the success data would be thrice each. Therefore, the total number of autonomous motion data collected was 300. Among these, 200~data and 100~data were used as training and validation data, respectively. Therefore, the new training dataset comprised 224~data and the validation dataset comprised 124~data. The number of trials was 594, the same as in Section~\ref{subsec:PRELIMINARY EXPERIMENT}. The number of training times of the NN was determined considering the validation loss, and 23000~times was adopted. This model is referred to as the self-supervised learning model~1. The loss graph for the self-supervised learning model~1 is illustrated in Fig.~\ref{fig:fig15}~(b)

\Figure[t!](topskip=0pt, botskip=0pt, midskip=0pt)[width=80mm]{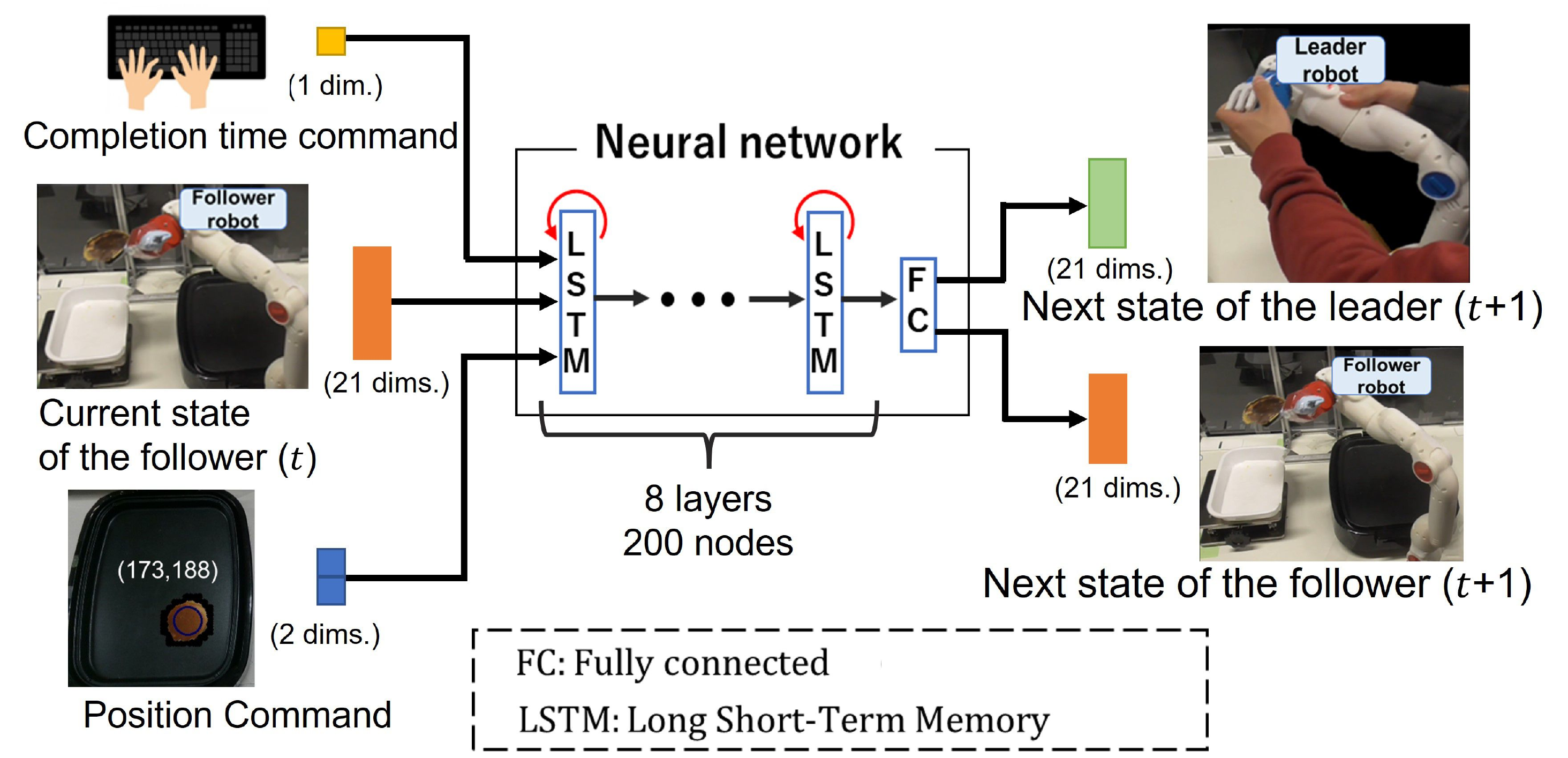}
{Structure of the NN implemented in this study \label{fig:fig10}}

\Figure[t!](topskip=0pt, botskip=0pt, midskip=0pt)[width=80mm]{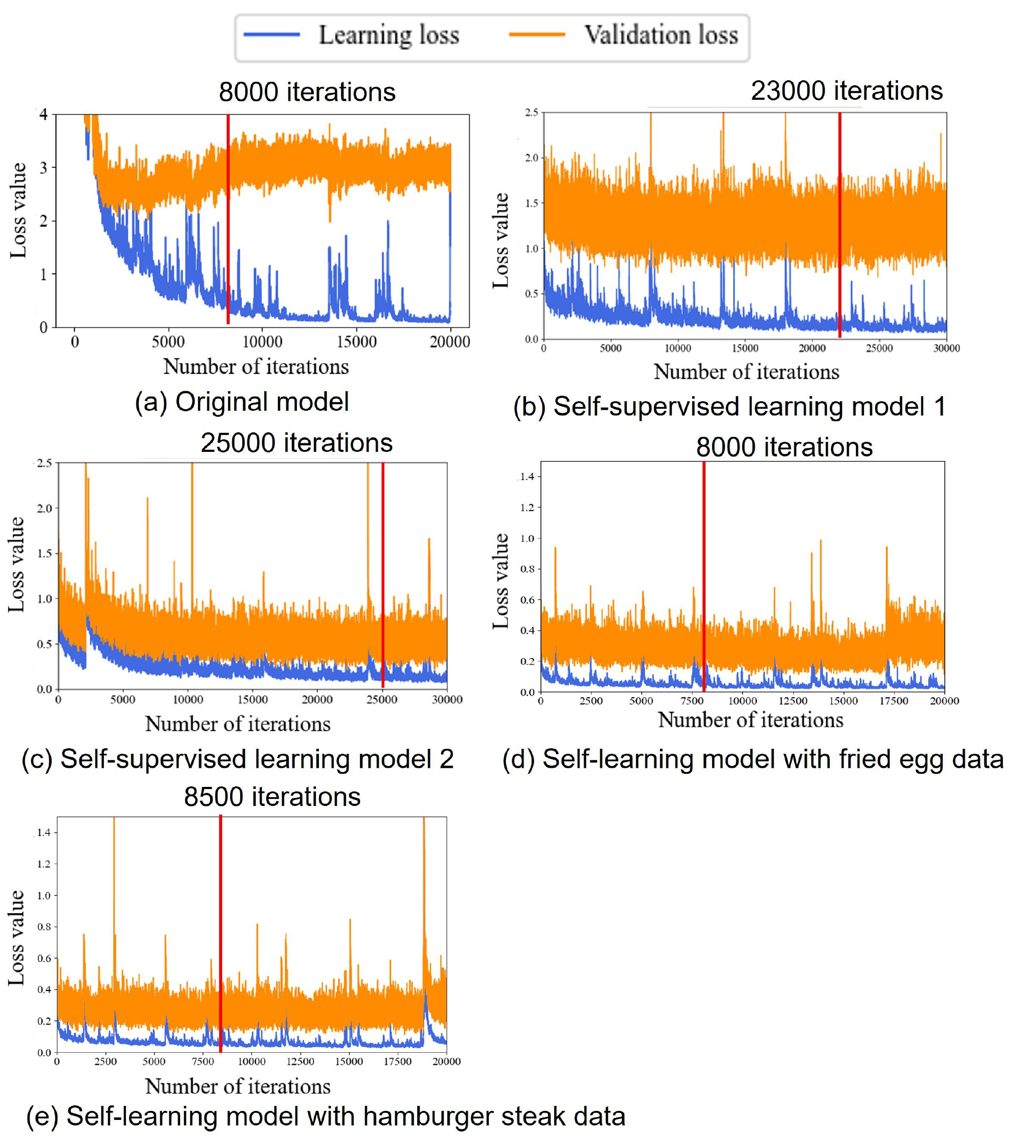}
{Training loss and validation loss for each neural network \label{fig:fig15}}


 \begin{figure*}[tb]
    \centering
        \includegraphics[width=155mm]{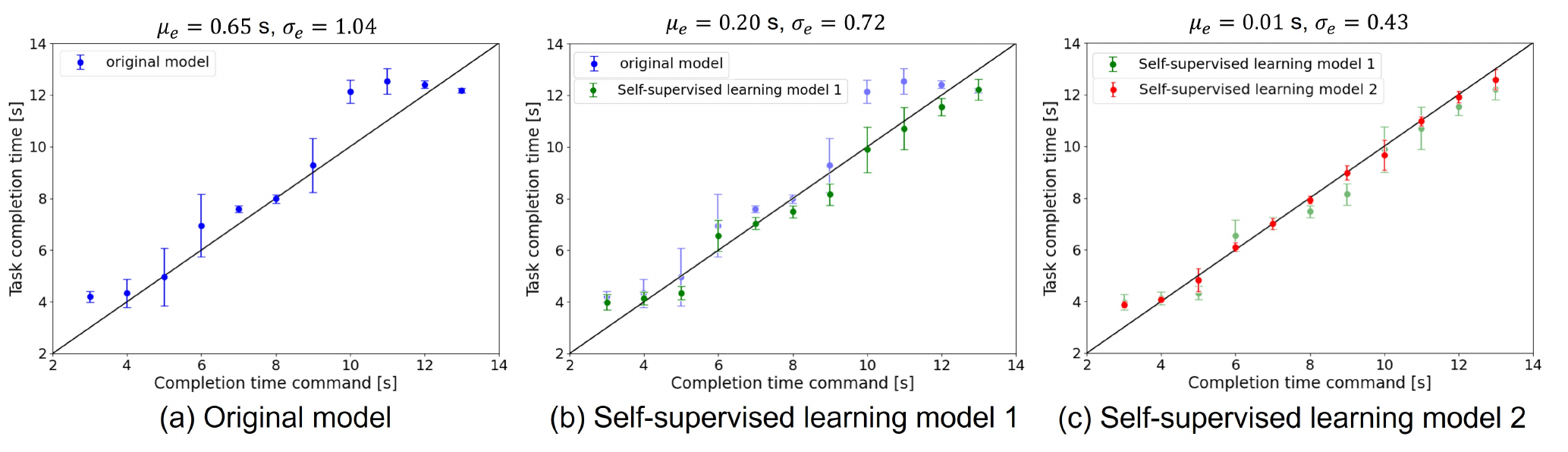}
        \caption{Comparison of the reproducibility of task completion time  using the proposed method. \newline In (b) and (c), the results of the model before fine tuning are plotted.}
        \label{fig:fig11}
\end{figure*}

\subsubsection{Result}
\label{subsubsec:Result2}
The task success rates for the self-supervised learning model~1 are listed in Table~\ref{tab:tab4}. We determined that the success rate of the self-supervised learning model~1 improved to 81.6\%, which was approximately double the success rate of the original model. With the increase in autonomous motion data, both spatial interpolation and dynamics learning progressed, allowing the robot to complete tasks at task completion times and positions that were unsuccessful in the original model. The major improvement in the spatial direction was the achievement of a task success rate of 56.1~\% in the upper, where the original model had a low success rate. In the temporal direction, the robot could rotate its wrist joint with an appropriate force and angular velocity, which reduced the number of failures when the pancake was dropped during the lifting phase. 

In the experimental results, there are only a few conditions in which the original model has a higher success rate than the self-supervised learning model 1. This is thought to be because the number of trials per condition was 6 times, which caused a variance in the success rate. However, the success rate for the overall speed is clearly lower for the faster speeds. This indicates that fast operation is more complex than slow operation. Future training with increased amounts of appropriate autonomous data will enable high speed operation.

For position commands, there was no difference in success rate between positions demonstrated by humans and positions not demonstrated by humans. However, for the task completion time command, the success rate was lower for the speed that was not demonstrated by a human compared to the speed demonstrated by a human. These results indicate that it is difficult to learn the dynamics of the change in speed from autonomous motion data, and it is necessary to collect more autonomous motion data at various speeds for learning.

The reproducibility of the task completion time of the self-supervised learning model~1 is illustrated in Fig.~\ref{fig:fig11}~(b). Here, as 2) in Section~\ref{subsec:PRELIMINARY EXPERIMENT}, only the results of successful trials are employed to calculate the mean and variance. Whereas the error of the original model was 0.65~s on average and 1.04 in variance, the error of the self-supervised learning model~1 was 0.2~s on average and 0.72 in variance, indicating that the reproducibility of task completion time was greatly improved. In particular, for the task completion time commands of 5, 6~s and 10, 11~s, which had large errors in the original model, the plot points are closer to the commanded values and improvement is remarkable. Although the original model tended to operate slowly in response to commands, this issue has been resolved.

\subsection{Self-supervised learning(2nd time)}
\label{subsec:Self-supervised learning(2nd time)}
\subsubsection{Description}
\label{subsubsec:Description3}
To test the case of collecting more autonomous motion data, an experiment was conducted using the data collected in Section~\ref{subsec:Self-supervised learning(1st time)}. For each of the 180~patterns that were successful at least once in Section~\ref{subsec:Self-supervised learning(1st time)}, we obtained additional data so that there were three successful data. The total number of autonomous action data was 540. Among these, 180~data and 360~data were used as training data and validation data, respectively. Thus, the total number of data for training was 404, and that for validation was 484. The total number of trials was 594, the same as in Sections~\ref{subsec:PRELIMINARY EXPERIMENT} and~\ref{subsec:Self-supervised learning(1st time)}. The number of training times of the NN was determined considering the validation loss, and 25000 times was used. This model is referred to as the self-supervised learning model~2. The loss graph for the self-supervised learning model~2 is illustrated in Fig.~\ref{fig:fig15}~(c).


\begin{table*}[]
\centering
\caption{Task success rate of the self-supervised learning model 1}
\scalebox{0.87}{ 
\begin{tabular}{c||ccccccccc||c}
\hline \hline
\multirow{3}{*}{\begin{tabular}[c]{@{}c@{}}Task completion\\ time command {[}s{]}\end{tabular}} & \multicolumn{9}{c|}{Position command}                                                                                                                                                                                                                                                                                                                                        & \multirow{3}{*}{\begin{tabular}[c]{@{}c@{}}Speed\\ total\end{tabular}} \\ \cline{2-10}
                                                                                                & \multicolumn{4}{c|}{Learned}                                                                                                                                               & \multicolumn{5}{c|}{Unlearned}                                                                                                                                                                  &                                                                        \\ \cline{2-10}
                                                                                                & \multicolumn{1}{c|}{Lower left}        & \multicolumn{1}{c|}{Upper left}           & \multicolumn{1}{c|}{Lower right}          & \multicolumn{1}{c|}{Upper right}          & \multicolumn{1}{c|}{Center}               & \multicolumn{1}{c|}{Left}              & \multicolumn{1}{c|}{Lower}                & \multicolumn{1}{c|}{Right}                & Upper              &                                                                        \\ \hline
3.00                                                                                            & \multicolumn{1}{c|}{0.0(0/6)}          & \multicolumn{1}{c|}{50.0(3/6)}            & \multicolumn{1}{c|}{\textbf{100(6/6)}}    & \multicolumn{1}{c|}{\textbf{83.3(5/6)}}   & \multicolumn{1}{c|}{50.0(3/6)}            & \multicolumn{1}{c|}{0.0(0/6)}          & \multicolumn{1}{c|}{\textbf{100(6/6)}}    & \multicolumn{1}{c|}{\textbf{100(6/6)}}    & 66.7(4/6)          & 61.1(33/54)                                                            \\ \hline
4.00                                                                                            & \multicolumn{1}{c|}{$\ast$\textbf{100(6/6)}} & \multicolumn{1}{c|}{$\ast$66.7(4/6)}            & \multicolumn{1}{c|}{$\ast$\textbf{100(6/6)}}    & \multicolumn{1}{c|}{$\ast$\textbf{100(6/6)}}    & \multicolumn{1}{c|}{\textbf{100(6/6)}}    & \multicolumn{1}{c|}{33.3(2/6)}         & \multicolumn{1}{c|}{\textbf{100(6/6)}}    & \multicolumn{1}{c|}{\textbf{100(6/6)}}    & \textbf{100(6/6)}  & \textbf{88.9(48/54)}                                                   \\ \hline
5.00                                                                                            & \multicolumn{1}{c|}{\textbf{100(6/6)}} & \multicolumn{1}{c|}{66.7(4/6)}            & \multicolumn{1}{c|}{\textbf{83.3(5/6)}}   & \multicolumn{1}{c|}{\textbf{100(6/6)}}    & \multicolumn{1}{c|}{\textbf{100(6/6)}}    & \multicolumn{1}{c|}{16.7(1/6)}         & \multicolumn{1}{c|}{\textbf{100(6/6)}}    & \multicolumn{1}{c|}{\textbf{83.3(5/6)}}   & 0.0(0/6)           & 72.2(39/54)                                                            \\ \hline
6.00                                                                                            & \multicolumn{1}{c|}{50.0(3/6)}         & \multicolumn{1}{c|}{16.7(1/6)}            & \multicolumn{1}{c|}{\textbf{100(6/6)}}    & \multicolumn{1}{c|}{\textbf{100(6/6)}}    & \multicolumn{1}{c|}{\textbf{100(6/6)}}    & \multicolumn{1}{c|}{\textbf{100(6/6)}} & \multicolumn{1}{c|}{\textbf{100(6/6)}}    & \multicolumn{1}{c|}{33.3(2/6)}            & 0.0(0/6)           & 66.7(36/54)                                                            \\ \hline
7.00                                                                                            & \multicolumn{1}{c|}{\textbf{100(6/6)}} & \multicolumn{1}{c|}{\textbf{100(6/6)}}    & \multicolumn{1}{c|}{\textbf{100(6/6)}}    & \multicolumn{1}{c|}{\textbf{100(6/6)}}    & \multicolumn{1}{c|}{66.7(4/6)}            & \multicolumn{1}{c|}{\textbf{100(6/6)}} & \multicolumn{1}{c|}{\textbf{100(6/6)}}    & \multicolumn{1}{c|}{\textbf{100(6/6)}}    & 16.7(1/6)          & \textbf{87.0(47/54)}                                                   \\ \hline
8.00                                                                                            & \multicolumn{1}{c|}{$\ast$\textbf{100(6/6)}} & \multicolumn{1}{c|}{$\ast$\textbf{100(6/6)}}    & \multicolumn{1}{c|}{$\ast$\textbf{100(6/6)}}    & \multicolumn{1}{c|}{$\ast$\textbf{100(6/6)}}    & \multicolumn{1}{c|}{\textbf{83.3(5/6)}}   & \multicolumn{1}{c|}{\textbf{100(6/6)}} & \multicolumn{1}{c|}{\textbf{100(6/6)}}    & \multicolumn{1}{c|}{\textbf{100(6/6)}}    & \textbf{83.3(5/6)} & \textbf{96.3(52/54)}                                                   \\ \hline
9.00                                                                                            & \multicolumn{1}{c|}{\textbf{100(6/6)}} & \multicolumn{1}{c|}{\textbf{100(6/6)}}    & \multicolumn{1}{c|}{\textbf{100(6/6)}}    & \multicolumn{1}{c|}{\textbf{100(6/6)}}    & \multicolumn{1}{c|}{33.3(2/6)}            & \multicolumn{1}{c|}{\textbf{100(6/6)}} & \multicolumn{1}{c|}{\textbf{100(6/6)}}    & \multicolumn{1}{c|}{\textbf{100(6/6)}}    & \textbf{100(6/6)}  & \textbf{92.6(50/54)}                                                   \\ \hline
10.0                                                                                            & \multicolumn{1}{c|}{\textbf{100(6/6)}} & \multicolumn{1}{c|}{\textbf{100(6/6)}}    & \multicolumn{1}{c|}{\textbf{100(6/6)}}    & \multicolumn{1}{c|}{\textbf{100(6/6)}}    & \multicolumn{1}{c|}{\textbf{100(6/6)}}    & \multicolumn{1}{c|}{\textbf{100(6/6)}} & \multicolumn{1}{c|}{50.0(3/6)}            & \multicolumn{1}{c|}{\textbf{100(6/6)}}    & \textbf{100(6/6)}  & \textbf{94.4(51/54)}                                                   \\ \hline
11.0                                                                                            & \multicolumn{1}{c|}{\textbf{100(6/6)}} & \multicolumn{1}{c|}{\textbf{100(6/6)}}    & \multicolumn{1}{c|}{\textbf{100(6/6)}}    & \multicolumn{1}{c|}{\textbf{100(6/6)}}    & \multicolumn{1}{c|}{66.7(4/6)}            & \multicolumn{1}{c|}{50.0(3/6)}         & \multicolumn{1}{c|}{0.0(0/6)}             & \multicolumn{1}{c|}{\textbf{100(6/6)}}    & \textbf{83.3(5/6)} & 77.8(42/54)                                                            \\ \hline
12.0                                                                                            & \multicolumn{1}{c|}{$\ast$\textbf{100(6/6)}} & \multicolumn{1}{c|}{$\ast$\textbf{100(6/6)}}    & \multicolumn{1}{c|}{$\ast$\textbf{100(6/6)}}    & \multicolumn{1}{c|}{$\ast$\textbf{100(6/6)}}    & \multicolumn{1}{c|}{\textbf{100(6/6)}}    & \multicolumn{1}{c|}{\textbf{100(6/6)}} & \multicolumn{1}{c|}{\textbf{100(6/6)}}    & \multicolumn{1}{c|}{\textbf{100(6/6)}}    & 33.3(2/6)          & \textbf{92.6(50/54)}                                                   \\ \hline
13.0                                                                                            & \multicolumn{1}{c|}{0.0(0/6)}          & \multicolumn{1}{c|}{\textbf{83.3(5/6)}}   & \multicolumn{1}{c|}{50.0(3/6)}            & \multicolumn{1}{c|}{\textbf{100(6/6)}}    & \multicolumn{1}{c|}{\textbf{100(6/6)}}    & \multicolumn{1}{c|}{\textbf{100(6/6)}} & \multicolumn{1}{c|}{50.0(3/6)}            & \multicolumn{1}{c|}{\textbf{100(6/6)}}    & 33.3(2/6)          & 68.5(37/54)                                                            \\ \hline \hline
\begin{tabular}[c]{@{}c@{}}Position\\ total\end{tabular}                                        & \multicolumn{1}{c|}{77.3(51/66)}       & \multicolumn{1}{c|}{\textbf{80.3(53/66)}} & \multicolumn{1}{c|}{\textbf{93.9(62/66)}} & \multicolumn{1}{c|}{\textbf{98.5(65/66)}} & \multicolumn{1}{c|}{\textbf{81.8(54/66)}} & \multicolumn{1}{c|}{72.7(48/66)}       & \multicolumn{1}{c|}{\textbf{81.8(54/66)}} & \multicolumn{1}{c|}{\textbf{92.4(61/66)}} & 56.1(37/66)        & \textbf{81.6(485/594)}                                                 \\ \hline \hline
\end{tabular}
}
\raggedright
$\ast$:Learned position and speed commands
\label{tab:tab4}
\end{table*}


\begin{table*}[]
\centering
\caption{Task success rate of the self-supervised learning model 2}
\scalebox{0.87}{ 
\begin{tabular}{c||ccccccccc||c}
\hline \hline
\multirow{3}{*}{\begin{tabular}[c]{@{}c@{}}Task completion\\ time command {[}s{]}\end{tabular}} & \multicolumn{9}{c|}{Position command}                                                                                                                                                                                                                                                                                                                                          & \multirow{3}{*}{\begin{tabular}[c]{@{}c@{}}Speed\\ total\end{tabular}} \\ \cline{2-10}
                                                                                                & \multicolumn{4}{c|}{Learned}                                                                                                                                              & \multicolumn{5}{c|}{Unlearned}                                                                                                                                                                     &                                                                        \\ \cline{2-10}
                                                                                                & \multicolumn{1}{c|}{Lower left}         & \multicolumn{1}{c|}{Upper left}           & \multicolumn{1}{c|}{Lower right}        & \multicolumn{1}{c|}{Upper right}          & \multicolumn{1}{c|}{Center}               & \multicolumn{1}{c|}{Left}                 & \multicolumn{1}{c|}{Lower}              & \multicolumn{1}{c|}{Right}                & Upper                &                                                                        \\ \hline
3.00                                                                                            & \multicolumn{1}{c|}{66.7(4/6)}          & \multicolumn{1}{c|}{\textbf{100(6/6)}}    & \multicolumn{1}{c|}{\textbf{83.3(5/6)}} & \multicolumn{1}{c|}{\textbf{100(6/6)}}    & \multicolumn{1}{c|}{\textbf{83.3(5/6)}}   & \multicolumn{1}{c|}{33.3(2/6)}            & \multicolumn{1}{c|}{\textbf{83.3(5/6)}} & \multicolumn{1}{c|}{50.0(3/6)}            & \textbf{100(6/6)}    & 77.8(42/54)                                                            \\ \hline
4.00                                                                                            & \multicolumn{1}{c|}{$\ast$\textbf{100(6/6)}}  & \multicolumn{1}{c|}{$\ast$\textbf{100(6/6)}}    & \multicolumn{1}{c|}{$\ast$\textbf{100(6/6)}}  & \multicolumn{1}{c|}{$\ast$\textbf{100(6/6)}}    & \multicolumn{1}{c|}{\textbf{83.3(5/6)}}   & \multicolumn{1}{c|}{66.7(4/6)}            & \multicolumn{1}{c|}{\textbf{83.3(5/6)}} & \multicolumn{1}{c|}{\textbf{100(6/6)}}    & \textbf{100(6/6)}    & \textbf{92.6(50/54)}                                                   \\ \hline
5.00                                                                                            & \multicolumn{1}{c|}{33.3(2/6)}          & \multicolumn{1}{c|}{\textbf{100(6/6)}}    & \multicolumn{1}{c|}{0.0(0/6)}           & \multicolumn{1}{c|}{66.7(4/6)}            & \multicolumn{1}{c|}{\textbf{83.3(5/6)}}   & \multicolumn{1}{c|}{\textbf{100(6/6)}}    & \multicolumn{1}{c|}{\textbf{100(6/6)}}  & \multicolumn{1}{c|}{\textbf{100(6/6)}}    & \textbf{83.3(5/6)}   & 74.1(40/54)                                                            \\ \hline
6.00                                                                                            & \multicolumn{1}{c|}{66.7(4/6)}          & \multicolumn{1}{c|}{\textbf{83.3(5/6)}}   & \multicolumn{1}{c|}{66.7(4/6)}          & \multicolumn{1}{c|}{\textbf{100(6/6)}}    & \multicolumn{1}{c|}{50.0(3/6)}            & \multicolumn{1}{c|}{\textbf{100(6/6)}}    & \multicolumn{1}{c|}{16.7(1/6)}          & \multicolumn{1}{c|}{\textbf{83.3(5/6)}}   & 16.7(1/6)            & 64.8(35/54)                                                            \\ \hline
7.00                                                                                            & \multicolumn{1}{c|}{66.7(4/6)}          & \multicolumn{1}{c|}{\textbf{100(6/6)}}    & \multicolumn{1}{c|}{\textbf{100(6/6)}}  & \multicolumn{1}{c|}{\textbf{100(6/6)}}    & \multicolumn{1}{c|}{\textbf{100(6/6)}}    & \multicolumn{1}{c|}{\textbf{100(6/6)}}    & \multicolumn{1}{c|}{\textbf{83.3(5/6)}} & \multicolumn{1}{c|}{\textbf{100(6/6)}}    & \textbf{100(6/6)}    & \textbf{94.4(51/54)}                                                   \\ \hline
8.00                                                                                            & \multicolumn{1}{c|}{$\ast$\textbf{100(6/6)}}  & \multicolumn{1}{c|}{$\ast$\textbf{100(6/6)}}    & \multicolumn{1}{c|}{$\ast$\textbf{100(6/6)}}  & \multicolumn{1}{c|}{$\ast$\textbf{100(6/6)}}    & \multicolumn{1}{c|}{\textbf{83.3(5/6)}}   & \multicolumn{1}{c|}{\textbf{100(6/6)}}    & \multicolumn{1}{c|}{\textbf{100(6/6)}}  & \multicolumn{1}{c|}{33.3(2/6)}            & 66.7(4/6)            & \textbf{87.0(47/54)}                                                   \\ \hline
9.00                                                                                            & \multicolumn{1}{c|}{\textbf{100(6/6)}}  & \multicolumn{1}{c|}{\textbf{83.3(5/6)}}   & \multicolumn{1}{c|}{\textbf{100(6/6)}}  & \multicolumn{1}{c|}{\textbf{100(6/6)}}    & \multicolumn{1}{c|}{66.7(4/6)}            & \multicolumn{1}{c|}{\textbf{100(6/6)}}    & \multicolumn{1}{c|}{\textbf{100(6/6)}}  & \multicolumn{1}{c|}{\textbf{100(6/6)}}    & \textbf{100(6/6)}    & \textbf{94.4(51/54)}                                                   \\ \hline
10.0                                                                                            & \multicolumn{1}{c|}{\textbf{100(6/6)}}  & \multicolumn{1}{c|}{\textbf{100(6/6)}}    & \multicolumn{1}{c|}{\textbf{83.3(5/6)}} & \multicolumn{1}{c|}{\textbf{100(6/6)}}    & \multicolumn{1}{c|}{\textbf{100(6/6)}}    & \multicolumn{1}{c|}{\textbf{83.3(5/6)}}   & \multicolumn{1}{c|}{66.7(4/6)}          & \multicolumn{1}{c|}{\textbf{100(6/6)}}    & \textbf{100(6/6)}    & \textbf{92.6(50/54)}                                                   \\ \hline
11.0                                                                                            & \multicolumn{1}{c|}{\textbf{100(6/6)}}  & \multicolumn{1}{c|}{\textbf{100(6/6)}}    & \multicolumn{1}{c|}{33.3(2/6)} & \multicolumn{1}{c|}{\textbf{100(6/6)}}    & \multicolumn{1}{c|}{\textbf{100(6/6)}}    & \multicolumn{1}{c|}{\textbf{100(6/6)}}    & \multicolumn{1}{c|}{\textbf{83.3(5/6)}} & \multicolumn{1}{c|}{\textbf{100(6/6)}}    & \textbf{100(6/6)}    & \textbf{90.7(49/54)}                                                   \\ \hline
12.0                                                                                            & \multicolumn{1}{c|}{$\ast$\textbf{83.3(5/6)}} & \multicolumn{1}{c|}{$\ast$\textbf{100(6/6)}}    & \multicolumn{1}{c|}{$\ast$\textbf{83.3(5/6)}} & \multicolumn{1}{c|}{$\ast$\textbf{100(6/6)}}    & \multicolumn{1}{c|}{\textbf{100(6/6)}}    & \multicolumn{1}{c|}{\textbf{100(6/6)}}    & \multicolumn{1}{c|}{\textbf{100(6/6)}}  & \multicolumn{1}{c|}{\textbf{100(6/6)}}    & 66.7(4/6)            & \textbf{92.6(50/54)}                                                   \\ \hline
13.0                                                                                            & \multicolumn{1}{c|}{33.3(2/6)}          & \multicolumn{1}{c|}{\textbf{100(6/6)}}    & \multicolumn{1}{c|}{\textbf{100(6/6)}}  & \multicolumn{1}{c|}{\textbf{100(6/6)}}    & \multicolumn{1}{c|}{\textbf{100(6/6)}}    & \multicolumn{1}{c|}{\textbf{100(6/6)}}    & \multicolumn{1}{c|}{50.0(3/6)}          & \multicolumn{1}{c|}{\textbf{100(6/6)}}    & 50.0(3/6)            & \textbf{81.5(44/54)}                                                   \\ \hline \hline
\begin{tabular}[c]{@{}c@{}}Position\\ total\end{tabular}                                        & \multicolumn{1}{c|}{77.3(51/66)}        & \multicolumn{1}{c|}{\textbf{97.0(64/66)}} & \multicolumn{1}{c|}{77.3(51/66)}        & \multicolumn{1}{c|}{\textbf{97.0(64/66)}} & \multicolumn{1}{c|}{\textbf{86.4(57/66)}} & \multicolumn{1}{c|}{\textbf{89.4(59/66)}} & \multicolumn{1}{c|}{78.8(52/66)}        & \multicolumn{1}{c|}{\textbf{87.9(58/66)}} & \textbf{80.3(53/66)} & \textbf{85.7(509/594)}                                                 \\ \hline \hline
\end{tabular}
}
\raggedright
$\ast$:Learned position and speed commands
\label{tab:tab5}
\end{table*}

\subsubsection{Result}
\label{subsubsec:Result3}
The task success rates for the self-supervised learning model~2 are presented in Table~\ref{tab:tab5}.
The overall task success rate was 85.7\%, which is a further improvement from the self-supervised learning model~1. The position commands bottom, lower right, lower and lower left were less successful than the other position commands, and there were many failures to put the turner on top of the pancake.
As anticipated in 2) of Section~\ref{subsec:Self-supervised learning(1st time)}, the increase in autonomous operation data improved the success rate at high speeds. This result indicates that learning the dynamics changing with speed progressed, and NN generated appropriate behaviors.

The reproducibility of the task completion time of the self-supervised learning model~2 is illustrated in Fig.~\ref{fig:fig11}~(c). Here, as 2) in Section~\ref{subsec:PRELIMINARY EXPERIMENT}, only the results of successful trials are used to calculate the mean and variance. The self-supervised learning model~2 exhibits very high reproducibility, with a mean error of 0.01~s and variance of 0.43 for the task completion time. In particular, the task completion time commands of 6, 10, and 11~s, for which the variance was large in the self-supervised learning model~1, indicated very small variance. In the self-supervised learning models~1 and 2, the number of operation data used to calculate the variance of 6, 10, and 11~s was 36, 51, 42~ and 36, 50, 49, indicating that the variability did not decrease with the increase in the number of successful data.
These results indicate that both temporal and spatial learning can be advanced using the proposed method, and variable speed nonprehensile manipulation can be performed considering the dynamics of the environment and object.

\Figure[t!](topskip=0pt, botskip=0pt, midskip=0pt)[width=60mm]{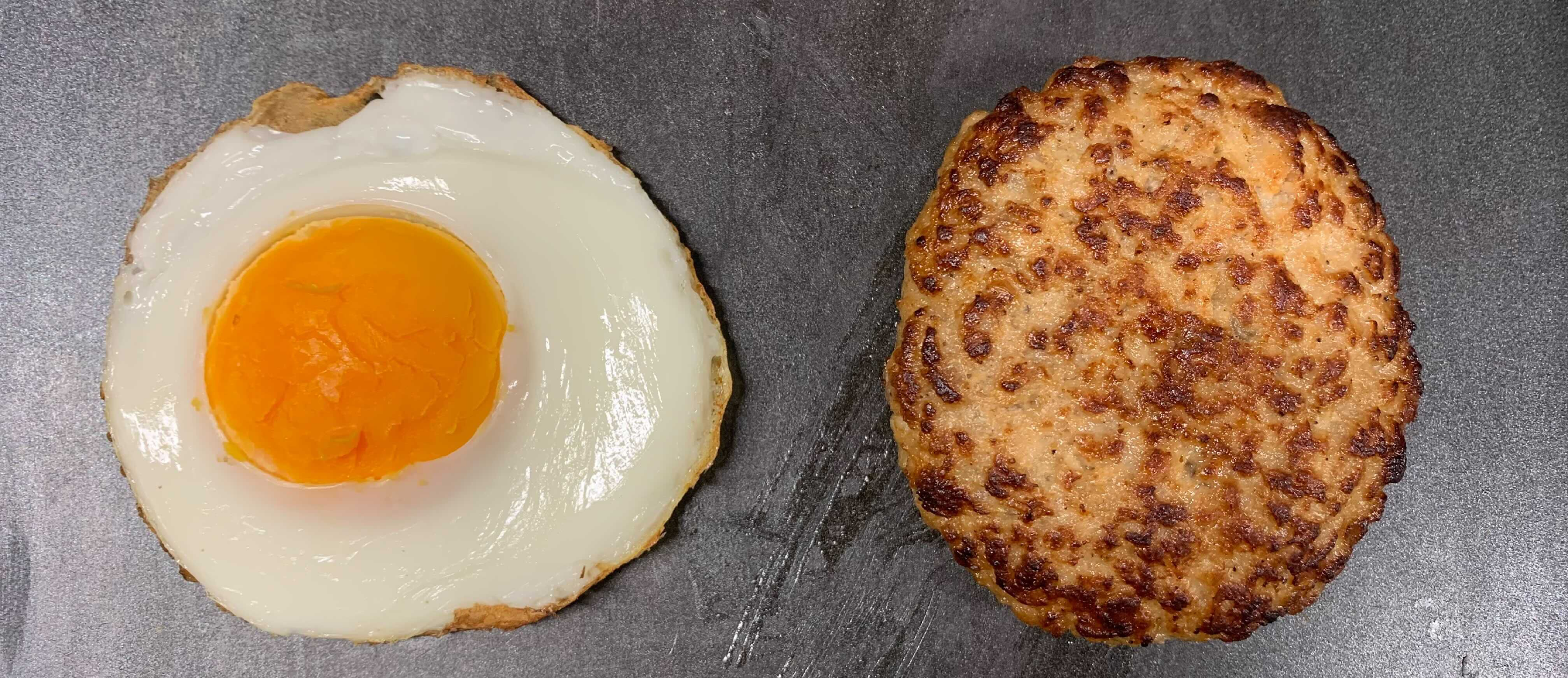}
{Unlearned objects. left;fried egg,right;hamburger-steak \label{fig:fig17}}

\begin{table}[]
\centering
\caption{Task success rate for untrained objects}
\scalebox{0.95}{ 
\begin{tabular}{c||c|c}
\hline 
Object          & Self-supervised learning model 2 & After self-learning \\ \hline \hline
Fried egg       & 72.1(214/297)                    & 78.8(234/297)       \\ \hline
Hamburger steak & 73.1(217/297)                    & 73.4(218/297)       \\ \hline
\end{tabular}
}
\label{tab:tab6}
\end{table}

\subsection{Evaluation of generalization performance for untrained objects}
\label{subsec:Evaluation of generalization performance for untrained objects}
\subsubsection{Description}
\label{subsubsec:Description4}
Using the self-supervised learning model~2, we evaluated the generalization performance for untrained objects. The untrained objects used in this study were the fried egg and the hamburger-steak shown in the Fig.~\ref{fig:fig17}, which weighed 42~g and 89~g, respectively. Three trials each were conducted under the same conditions of position and task completion time as in Section~\ref{subsec:PRELIMINARY EXPERIMENT}. Therefore, 297~trials (9~[position] × 11~[completion time] × 3~[trial]) were conducted for each object. 
Furthermore, we examined whether fine-tuning the self-supervised learning model~2 with the autonomous action data performed on each object would improve the success rate of the task by learning the physical properties of untrained objects. In this fine-tuning, only the motion data of a specific object was used, not the pancake data.

\subsubsection{Result}
\label{subsubsec:Result4}
As shown on the left side of Table~\ref{tab:tab6}, the task success rates for self-supervised learning model~2 were 72.1\% for the fried egg and 73.1\% for the hamburger-steak. Because the mass, friction of the fried egg or the hamburger-steak are very different from those of a pancake, it is difficult for them to succeed in the task. However, the self-supervised learning model~2, which had been sufficiently trained in nonprehensile manipulation, could successfully perform the task with a high success rate of more than 70\% for unlearned objects.

Furthermore, the task success rates of the NN after fine-tuning with the fried egg and hamburger-steak autonomous behavior data is also shown on the right side of Table~\ref{tab:tab6}. The numbers of times the NNs were trained was 8000 for the fried egg task and 8500 for the hamburger-steak task, and the loss graphs are shown in Figs.~\ref{fig:fig15}~(d),~(e). 
The success rate of the self-learning model with the fried egg was 78.8\%, and the success rate of the self-learning model with the hamburger-steak was 73.4\%, both of which were slightly better than the results before self-learning. However, the success rate was below 80\%, and the improvement in task success rate was less significant when compared to the self-learning of pancakes.  
Therefore, under the conditions of this study, it was possible to improve the task success rate of nonprehensile manipulation by using the proposed method, though the growth rate was shown to decrease as learning progressed.

\subsection{Discussion}
\label{subsec:Discussion}
\begin{figure*}[tb]
    \centering
        \includegraphics[width=160mm]{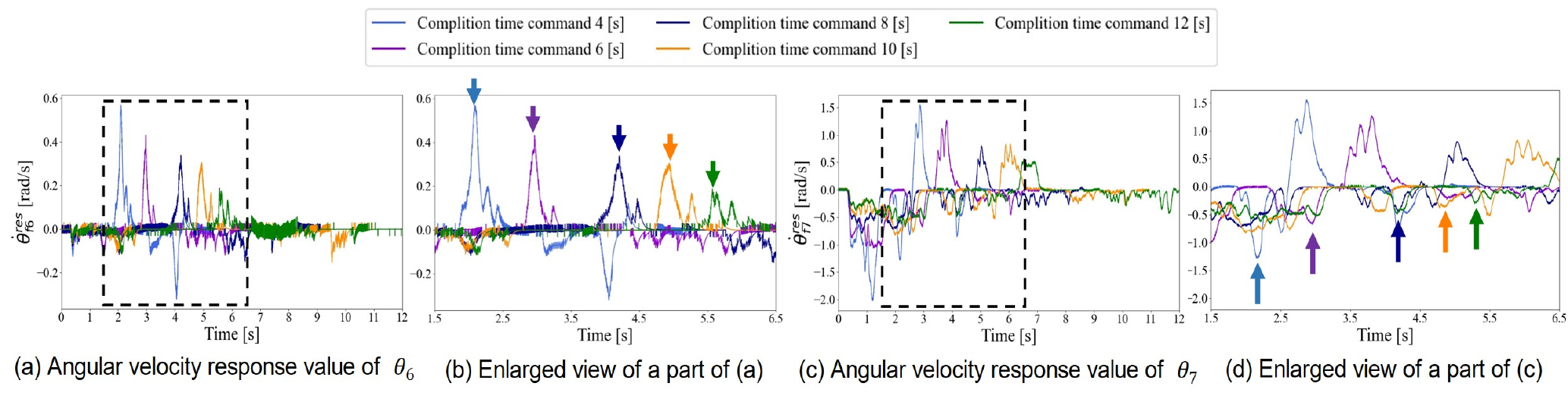}
        \caption{Comparison of angular velocity response values when executing a task at each speed. The dashed lines in (a) and (c) represent the duration of the transporting and lifting phases, respectively. (b) and (d) focus on the time of the dashed boxes in (a) and (c), respectively.}
        \label{fig:fig14}
\end{figure*}

Fig.~\ref{fig:fig14} illustrates the angular velocity response values of $\theta_6,\theta_7$ when the task is performed at multiple speeds using the self-supervised learning model~2. We performed the task in the posture illustrated in Fig.~\ref{fig:fig16}. From Fig.~\ref{fig:fig14}~(a), it can be confirmed that the wrist joints of $\theta_6$ are rotated rapidly as the task completion time becomes shorter. $\theta_6$ corresponds to the snap of the wrist that moves the turner horizontally, and high-speed rotation of the wrist prevents the pancake from flying away owing to  centrifugal force during the transporting phase. It was confirmed that the NN learns the inertia caused by the fast transportation motion. In the angular velocity response values for $\theta_7$ illustrated in Fig.~\ref{fig:fig14}~(b), $\theta_7$ corresponds to the snap of the wrist that moves the turner up and down. Fig.~\ref{fig:fig14}~(b) illustrates that the angular velocity in the lifting phase changes according to the task completion time, and that the NN generates appropriate motions to prevent the pancake from slipping off and being thrown away in the lifting phase.

From the above data, it is confirmed that the proposed method can learn the environment, pancake, and dynamics between them from the autonomous operation data. Moreover, the results of 2) in Section~\ref{subsec:Self-supervised learning(1st time)} indicate that providing various conditions in the spatio-temporal direction of interpolation could learn efficient self-learning. Therefore, we speculate  that it is important to train NNs considering both time and space to efficiently learn the dynamic behavior of robots.



\Figure[t!](topskip=0pt, botskip=0pt, midskip=0pt)[width=80mm]{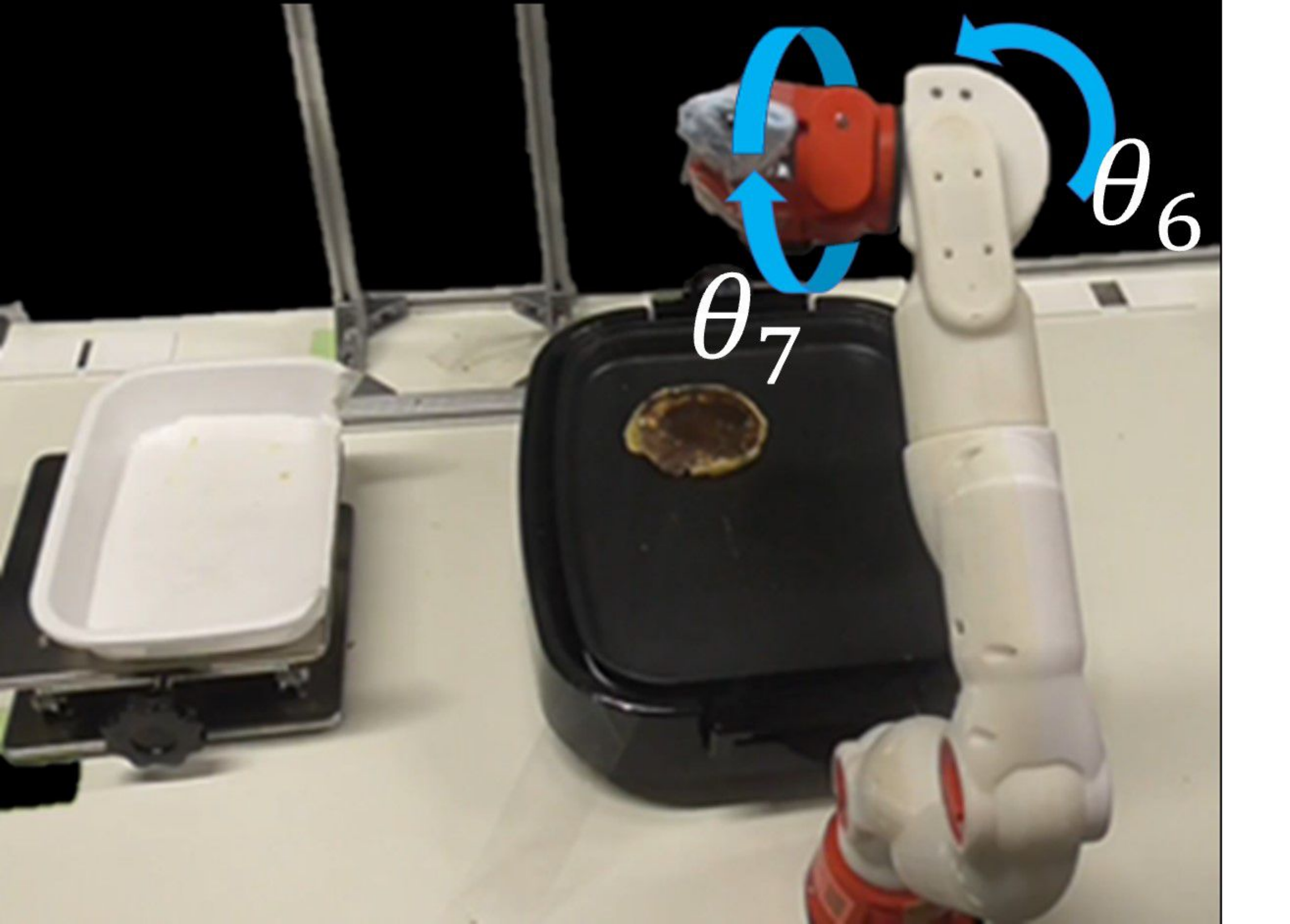}
{State of each joint of $\theta_6$ and $\theta_7$ during task execution \label{fig:fig16}}


The experimental results indicated that the self-supervised learning model~2 was able to reproduce the speed with very high accuracy, with an error of 0.01~s on average and a variance of 0.43. However, the error of 48~ datasets collected by humans at 4, 8, and 12~s had a mean of 0.17~s and variance of 0.46. Hence, self-supervised learning model~2 can perform the task with higher accuracy than humans. In conventional imitation learning, all demonstrations have to be taught at approximately the same speed of movement to stabilize learning. In fact, in our conventional method~\cite{ref29}, a human demonstrated the task while listening to the sound of a metronome, so that the task execution time would be the same; however, this was very laborious, and it was also impossible to reproduce demonstrations with the same motion speeds. On the contrary, using the proposed method, accurate speed labels can be generated later, and motion speed can be adjusted from the autonomous motion data obtained from the trained NN. Hence, humans are freed from collecting data at the exact rate, and the model itself can be fitted to the exact speed through self-learning.

The proposed method can learn from all the time-series data of successful actions, whereas reinforcement learning learns from rewards determined only by the final states. Therefore, the proposed method is more sample efficient than reinforcement learning. Moreover, because imitation learning limits the range of action by human demonstration, the generated actions are relatively safe and can be subtly adjusted with similar autonomous action data. Reinforcement learning struggles with setting up a reward function that corresponds to task-related parameters, such as arbitrary motion speed, and is prone to overlearning for a specific motion speed.


In the experimental results of Section~\ref{subsec:Evaluation of generalization performance for untrained objects}, self-learning for unlearned objects did not significantly improve the success rate. We consider that this result is caused by the fact that the self-supervised learning model~2 has already overlearned for the pancake. In particular, in the structure of the NN used in this study, information about the physical properties of the manipulated object was not input. This means that the task is being executed with information such as shape and size unknown. Therefore, when a command with the same position and task completion time was input, the same motion with very little variation was generated without considering the size and shape of the object. Recently, methods using raw images have been studied ~\cite{ref54,ref56}. These studies add raw images to the NN input and learn to generate appropriate motions in response to changes in the position, and shape of the object. Therefore, we consider that the proposed method will be able to learn motions that consider various sizes, and shape by adding raw image data to the learning process. It may improve the success rate of 80\% or more for new objects. However, the input of images makes learning more difficult. In this case, curriculum learning~\cite{ref101,ref102}, which improves the prediction accuracy of the model by initially learning with data under simple conditions and gradually making the model more difficult, could be introduced to solve the problem. The combination of these methods and the proposed method is expected to be effective for untrained objects, because it considers the shape and size of the manipulated object and avoids overlearning for a specific object.

As mentioned in the introduction, the field of machine learning has focused on geometric problems and has insufficiently considered temporal information such as speed. However, because several physical phenomena have been formulated and understood using differential equations, time information is an important factor in dynamic object manipulation. In the proposed method, the dynamics to be considered in nonprehensile manipulation can be learned by self-learning using a time index. In addition, the proposed method can be widely employed for several tasks using the task completion time as the time index. Therefore, our method is highly compatible with conventional self-supervised learning focusing on geometric problems, and when combined with conventional methods, it is expected to enable more complex and dynamic behaviors, depending on the environment and situation.
In recent years, research has been conducted to understand concepts such as color and shape in language and images from the physicality of the robot using self-supervised learning~\cite{ref55}. Further development of the proposed method may lead to the understanding of complex and abstract concepts correlated with speed, such as “fast or slow,” “strong or weak,” and “heavy or light,”  through the dynamic body movements of the robot.

\section{Conclusion}
\label{sec:Conclusion}
We proposed a self-supervised learning method that considers speed. In the proposed method, the NN was fine-tuned using only successful actions among the autonomous action data generated by the trained NN. Using the proposed method was able to improve the success rate in both spatio-temporal conditions. Although nonprehensile manipulation requires considering the dynamics between the environment and object, and it is difficult to perform the task at multiple speeds, the proposed method was feasible with as few as 24 pieces of supervised data. Furthermore, the proposed method was able to complete the task in a more accurate time than the given training data. After self-learning, the NN appropriately altered its force and trajectory according to the task completion time, confirming that it is capable of learning the dynamics between itself, the environment, manipulated objects, and these from automatically generated autonomous motion data.

In the past, self-supervised learning focused on spatial information and did not fully utilize temporal information. The proposed method is compatible with these conventional methods, expanding the possibilities of self-supervised learning and contributing greatly to the understanding of dynamic phenomena in robotic tasks.
However, the spatial information in this study is limited to the center position of the pancake at the beginning of the task, and the shape and size of the object were not considered. Therefore, our future work is to integrate the proposed method with a real-time image-based motion generation method~\cite{ref54} and a method that considers the shape and size of multiple objects~\cite{ref56} to expand the tasks that can be performed by the robot in space and time.

\section*{Appendix}
\label{sec:Appendix}
The experimental results for the 30~g and 90~g individual pancakes are presented Tables~\ref{tab:tab7}--\ref{tab:tab12}. As shown in the Tables~\ref{tab:tab7}--\ref{tab:tab12}, task success rates for each model did not differ significantly with pancake mass.



\begin{table*}[]
\caption{Task success rate of the original model for pancake of 30g}
\scalebox{0.88}{ 
\begin{tabular}{c|ccccccccc|c}
\hline
\multirow{3}{*}{\begin{tabular}[c]{@{}c@{}}Task completion\\ time command {[}s{]}\end{tabular}} & \multicolumn{9}{c|}{Position command}                                                                                                                                                                                                                                                                                                      & \multirow{3}{*}{\begin{tabular}[c]{@{}c@{}}Speed\\ total\end{tabular}} \\ \cline{2-10}
                                                                                                & \multicolumn{4}{c|}{Learned}                                                                                                                               & \multicolumn{5}{c|}{Unlearned}                                                                                                                                                &                                                                        \\ \cline{2-10}
                                                                                                & \multicolumn{1}{c|}{Lower left} & \multicolumn{1}{c|}{Upper left}        & \multicolumn{1}{c|}{Lower right}       & \multicolumn{1}{c|}{Upper right}       & \multicolumn{1}{c|}{Center}            & \multicolumn{1}{c|}{Left}              & \multicolumn{1}{c|}{Lower}             & \multicolumn{1}{c|}{Right}             & Upper     &                                                                        \\ \hline
3.00                                                                                            & \multicolumn{1}{c|}{0.0(0/3)}   & \multicolumn{1}{c|}{\textbf{100(3/3)}} & \multicolumn{1}{c|}{0.0(0/3)}          & \multicolumn{1}{c|}{\textbf{100(3/3)}} & \multicolumn{1}{c|}{0.0(0/3)}          & \multicolumn{1}{c|}{0.0(0/3)}          & \multicolumn{1}{c|}{\textbf{100(3/3)}} & \multicolumn{1}{c|}{0.0(0/3)}          & 0.0(0/3)  & 33.3(9/27)                                                             \\ \hline
4.00                                                                                            & \multicolumn{1}{c|}{33.3(1/3)}  & \multicolumn{1}{c|}{\textbf{100(3/3)}} & \multicolumn{1}{c|}{\textbf{100(3/3)}} & \multicolumn{1}{c|}{33.3(1/3)}         & \multicolumn{1}{c|}{\textbf{100(3/3)}} & \multicolumn{1}{c|}{\textbf{100(3/3)}} & \multicolumn{1}{c|}{\textbf{100(3/3)}} & \multicolumn{1}{c|}{0.0(0/3)}          & 0.0(0/3)  & 63.0(17/27)                                                            \\ \hline
5.00                                                                                            & \multicolumn{1}{c|}{0.0(0/3)}   & \multicolumn{1}{c|}{\textbf{100(3/3)}} & \multicolumn{1}{c|}{0.0(0/3)}          & \multicolumn{1}{c|}{\textbf{100(3/3)}} & \multicolumn{1}{c|}{\textbf{100(3/3)}} & \multicolumn{1}{c|}{\textbf{100(3/3)}} & \multicolumn{1}{c|}{33.3(1/3)}         & \multicolumn{1}{c|}{33.3(1/3)}         & 0.0(0/3)  & 51.9(14/27)                                                            \\ \hline
6.00                                                                                            & \multicolumn{1}{c|}{0.0(0/3)}   & \multicolumn{1}{c|}{0.0(0/3)}          & \multicolumn{1}{c|}{0.0(0/3)}          & \multicolumn{1}{c|}{\textbf{100(3/3)}} & \multicolumn{1}{c|}{0.0(0/3)}          & \multicolumn{1}{c|}{\textbf{100(3/3)}} & \multicolumn{1}{c|}{\textbf{100(3/3)}} & \multicolumn{1}{c|}{0.0(0/3)}          & 0.0(0/3)  & 33.3(9/27)                                                             \\ \hline
7.00                                                                                            & \multicolumn{1}{c|}{33.3(1/3)}  & \multicolumn{1}{c|}{0.0(0/3)}          & \multicolumn{1}{c|}{0.0(0/3)}          & \multicolumn{1}{c|}{\textbf{100(3/3)}} & \multicolumn{1}{c|}{0.0(0/3)}          & \multicolumn{1}{c|}{0.0(0/3)}          & \multicolumn{1}{c|}{0.0(0/3)}          & \multicolumn{1}{c|}{66.7(2/3)}         & 0.0(0/3)  & 22.2(6/27)                                                             \\ \hline
8.00                                                                                            & \multicolumn{1}{c|}{66.7(2/3)}  & \multicolumn{1}{c|}{0.0(0/3)}          & \multicolumn{1}{c|}{66.7(2/3)}         & \multicolumn{1}{c|}{\textbf{100(3/3)}} & \multicolumn{1}{c|}{0.0(0/3)}          & \multicolumn{1}{c|}{0.0(0/3)}          & \multicolumn{1}{c|}{33.3(1/3)}         & \multicolumn{1}{c|}{33.3(1/3)}         & 0.0(0/3)  & 33.3(9/27)                                                             \\ \hline
9.00                                                                                            & \multicolumn{1}{c|}{66.7(2/3)}  & \multicolumn{1}{c|}{0.0(0/3)}          & \multicolumn{1}{c|}{33.3(1/3)}         & \multicolumn{1}{c|}{\textbf{100(3/3)}} & \multicolumn{1}{c|}{0.0(0/3)}          & \multicolumn{1}{c|}{66.7(2/3)}         & \multicolumn{1}{c|}{0.0(0/3)}          & \multicolumn{1}{c|}{66.7(2/3)}         & 0.0(0/3)  & 37.0(10/27)                                                            \\ \hline
10.0                                                                                            & \multicolumn{1}{c|}{0.0(0/3)}   & \multicolumn{1}{c|}{0.0(0/3)}          & \multicolumn{1}{c|}{0.0(0/3)}          & \multicolumn{1}{c|}{\textbf{100(3/3)}} & \multicolumn{1}{c|}{66.7(2/3)}         & \multicolumn{1}{c|}{33.3(1/3)}         & \multicolumn{1}{c|}{33.3(1/3)}         & \multicolumn{1}{c|}{\textbf{100(3/3)}} & 0.0(0/3)  & 37.0(10/27)                                                            \\ \hline
11.0                                                                                            & \multicolumn{1}{c|}{0.0(0/3)}   & \multicolumn{1}{c|}{\textbf{100(3/3)}} & \multicolumn{1}{c|}{66.7(2/3)}         & \multicolumn{1}{c|}{66.7(2/3)}         & \multicolumn{1}{c|}{66.7(2/3)}         & \multicolumn{1}{c|}{0.0(0/3)}          & \multicolumn{1}{c|}{66.7(2/3)}         & \multicolumn{1}{c|}{33.3(1/3)}         & 33.3(1/3) & 48.1(13/27)                                                            \\ \hline
12.0                                                                                            & \multicolumn{1}{c|}{0.0(0/3)}   & \multicolumn{1}{c|}{\textbf{100(3/3)}} & \multicolumn{1}{c|}{0.0(0/3)}          & \multicolumn{1}{c|}{33.3(1/3)}         & \multicolumn{1}{c|}{\textbf{100(3/3)}} & \multicolumn{1}{c|}{0.0(0/3)}          & \multicolumn{1}{c|}{\textbf{100(3/3)}} & \multicolumn{1}{c|}{66.7(2/3)}         & 0.0(0/3)  & 44.4(12/27)                                                            \\ \hline
13.0                                                                                            & \multicolumn{1}{c|}{0.0(0/3)}   & \multicolumn{1}{c|}{\textbf{100(3/3)}} & \multicolumn{1}{c|}{0.0(0/3)}          & \multicolumn{1}{c|}{66.7(2/3)}         & \multicolumn{1}{c|}{0.0(0/3)}          & \multicolumn{1}{c|}{0.0(0/3)}          & \multicolumn{1}{c|}{0.0(0/3)}          & \multicolumn{1}{c|}{66.7(2/3)}         & 0.0(0/3)  & 25.9(7/27)                                                             \\ \hline
\begin{tabular}[c]{@{}c@{}}Position \\ total\end{tabular}                                       & \multicolumn{1}{c|}{18.2(6/33)} & \multicolumn{1}{c|}{54.5(18/33)}       & \multicolumn{1}{c|}{24.2(8/33)}        & \multicolumn{1}{c|}{81.8(27/33)}       & \multicolumn{1}{c|}{39.4(13/33)}       & \multicolumn{1}{c|}{36.4(12/33)}       & \multicolumn{1}{c|}{51.5(17/33)}       & \multicolumn{1}{c|}{42.4(14/33)}       & 3.0(1/33) & 39.1(116/297)                                                          \\ \hline
\end{tabular}
}
\label{tab:tab7}
\end{table*}

\begin{table*}[]
\caption{Task success rate of the original model for pancake of 90g}
\scalebox{0.88}{ 
\begin{tabular}{c|ccccccccc|c}
\hline
\multirow{3}{*}{\begin{tabular}[c]{@{}c@{}}Task completion\\ time command {[}s{]}\end{tabular}} & \multicolumn{9}{c|}{Position command}                                                                                                                                                                                                                                                                                                                & \multirow{3}{*}{\begin{tabular}[c]{@{}c@{}}Speed\\ total\end{tabular}} \\ \cline{2-10}
                                                                                                & \multicolumn{4}{c|}{Learned}                                                                                                                                      & \multicolumn{5}{c|}{Unlearned}                                                                                                                                                   &                                                                        \\ \cline{2-10}
                                                                                                & \multicolumn{1}{c|}{Lower left}        & \multicolumn{1}{c|}{Upper left}        & \multicolumn{1}{c|}{Lower right}       & \multicolumn{1}{c|}{Upper right}       & \multicolumn{1}{c|}{Center}             & \multicolumn{1}{c|}{Left}               & \multicolumn{1}{c|}{Lower}              & \multicolumn{1}{c|}{Right}             & Upper     &                                                                        \\ \hline
3.00                                                                                            & \multicolumn{1}{c|}{0.0(0/3)}          & \multicolumn{1}{c|}{\textbf{0.0(0/3)}} & \multicolumn{1}{c|}{0.0(0/3)}          & \multicolumn{1}{c|}{\textbf{100(3/3)}} & \multicolumn{1}{c|}{0.0(0/3)}           & \multicolumn{1}{c|}{0.0(0/3)}           & \multicolumn{1}{c|}{\textbf{100(3/3)}}  & \multicolumn{1}{c|}{0.0(0/3)}          & 0.0(0/3)  & 22.2(6/27)                                                             \\ \hline
4.00                                                                                            & \multicolumn{1}{c|}{0.0(0/3)}          & \multicolumn{1}{c|}{\textbf{100(3/3)}} & \multicolumn{1}{c|}{\textbf{0.0(0/3)}} & \multicolumn{1}{c|}{\textbf{100(3/3)}} & \multicolumn{1}{c|}{\textbf{100(3/3)}}  & \multicolumn{1}{c|}{\textbf{100(3/3)}}  & \multicolumn{1}{c|}{\textbf{100(3/3)}}  & \multicolumn{1}{c|}{0.0(0/3)}          & 0.0(0/3)  & 55.6(15/27)                                                            \\ \hline
5.00                                                                                            & \multicolumn{1}{c|}{\textbf{100(3/3)}} & \multicolumn{1}{c|}{\textbf{0.0(0/3)}} & \multicolumn{1}{c|}{0.0(0/3)}          & \multicolumn{1}{c|}{\textbf{100(3/3)}} & \multicolumn{1}{c|}{\textbf{100(3/3)}}  & \multicolumn{1}{c|}{\textbf{33.3(1/3)}} & \multicolumn{1}{c|}{\textbf{100(3/3)}}  & \multicolumn{1}{c|}{0.0(0/3)}          & 0.0(0/3)  & 48.1(13/27)                                                            \\ \hline
6.00                                                                                            & \multicolumn{1}{c|}{0.0(0/3)}          & \multicolumn{1}{c|}{0.0(0/3)}          & \multicolumn{1}{c|}{0.0(0/3)}          & \multicolumn{1}{c|}{\textbf{100(3/3)}} & \multicolumn{1}{c|}{0.0(0/3)}           & \multicolumn{1}{c|}{\textbf{0.0(0/3)}}  & \multicolumn{1}{c|}{\textbf{33.3(1/3)}} & \multicolumn{1}{c|}{33.3(1/3)}         & 0.0(0/3)  & 18.5(5/27)                                                             \\ \hline
7.00                                                                                            & \multicolumn{1}{c|}{\textbf{100(3/3)}} & \multicolumn{1}{c|}{0.0(0/3)}          & \multicolumn{1}{c|}{0.0(0/3)}          & \multicolumn{1}{c|}{\textbf{100(3/3)}} & \multicolumn{1}{c|}{0.0(0/3)}           & \multicolumn{1}{c|}{0.0(0/3)}           & \multicolumn{1}{c|}{0.0(0/3)}           & \multicolumn{1}{c|}{66.7(2/3)}         & 0.0(0/3)  & 29.6(8/27)                                                             \\ \hline
8.00                                                                                            & \multicolumn{1}{c|}{\textbf{100(3/3)}} & \multicolumn{1}{c|}{0.0(0/3)}          & \multicolumn{1}{c|}{\textbf{100(3/3)}} & \multicolumn{1}{c|}{\textbf{100(3/3)}} & \multicolumn{1}{c|}{0.0(0/3)}           & \multicolumn{1}{c|}{0.0(0/3)}           & \multicolumn{1}{c|}{\textbf{100(3/3)}}  & \multicolumn{1}{c|}{0.0(0/3)}          & 0.0(0/3)  & 44.4(12/27)                                                            \\ \hline
9.00                                                                                            & \multicolumn{1}{c|}{33.3(1/3)}         & \multicolumn{1}{c|}{0.0(0/3)}          & \multicolumn{1}{c|}{\textbf{100(3/3)}} & \multicolumn{1}{c|}{\textbf{100(3/3)}} & \multicolumn{1}{c|}{0.0(0/3)}           & \multicolumn{1}{c|}{33.3(1/3)}          & \multicolumn{1}{c|}{0.0(0/3)}           & \multicolumn{1}{c|}{\textbf{100(3/3)}} & 0.0(0/3)  & 40.7(11/27)                                                            \\ \hline
10.0                                                                                            & \multicolumn{1}{c|}{33.3(1/3)}         & \multicolumn{1}{c|}{0.0(0/3)}          & \multicolumn{1}{c|}{\textbf{100(3/3)}} & \multicolumn{1}{c|}{\textbf{100(3/3)}} & \multicolumn{1}{c|}{\textbf{100(3/3)}}  & \multicolumn{1}{c|}{0.0(0/3)}           & \multicolumn{1}{c|}{\textbf{100(3/3)}}  & \multicolumn{1}{c|}{\textbf{100(3/3)}} & 0.0(0/3)  & 59.3(16/27)                                                            \\ \hline
11.0                                                                                            & \multicolumn{1}{c|}{0.0(0/3)}          & \multicolumn{1}{c|}{\textbf{100(3/3)}} & \multicolumn{1}{c|}{\textbf{100(3/3)}} & \multicolumn{1}{c|}{\textbf{100(3/3)}} & \multicolumn{1}{c|}{\textbf{100(3/3)}}  & \multicolumn{1}{c|}{0.0(0/3)}           & \multicolumn{1}{c|}{\textbf{100(3/3)}}  & \multicolumn{1}{c|}{66.7(2/3)}         & 33.3(1/3) & 66.7(18/27)                                                            \\ \hline
12.0                                                                                            & \multicolumn{1}{c|}{0.0(0/3)}          & \multicolumn{1}{c|}{\textbf{100(3/3)}} & \multicolumn{1}{c|}{0.0(0/3)}          & \multicolumn{1}{c|}{\textbf{100(3/3)}} & \multicolumn{1}{c|}{\textbf{66.7(2/3)}} & \multicolumn{1}{c|}{0.0(0/3)}           & \multicolumn{1}{c|}{\textbf{33.3(1/3)}} & \multicolumn{1}{c|}{66.7(2/3)}         & 0.0(0/3)  & 40.7(11/27)                                                            \\ \hline
13.0                                                                                            & \multicolumn{1}{c|}{0.0(0/3)}          & \multicolumn{1}{c|}{\textbf{100(3/3)}} & \multicolumn{1}{c|}{0.0(0/3)}          & \multicolumn{1}{c|}{\textbf{100(3/3)}} & \multicolumn{1}{c|}{0.0(0/3)}           & \multicolumn{1}{c|}{0.0(0/3)}           & \multicolumn{1}{c|}{0.0(0/3)}           & \multicolumn{1}{c|}{66.7(2/3)}         & 0.0(0/3)  & 29.6(8/27)                                                             \\ \hline
\begin{tabular}[c]{@{}c@{}}Position \\ total\end{tabular}                                       & \multicolumn{1}{c|}{33.3(11/33)}       & \multicolumn{1}{c|}{36.4(12/33)}       & \multicolumn{1}{c|}{36.4(12/33)}       & \multicolumn{1}{c|}{100(33/33)}        & \multicolumn{1}{c|}{42.4(14/33)}        & \multicolumn{1}{c|}{15.2(5/33)}         & \multicolumn{1}{c|}{60.6(20/33)}        & \multicolumn{1}{c|}{45.5(15/33)}       & 3.0(1/33) & 41.4(123/297)                                                          \\ \hline
\end{tabular}
}
\label{tab:tab8}
\end{table*}


\begin{table*}[]
\caption{Task success rate of the self-supervised learning model 1 for pancake of 30g}
\scalebox{0.88}{ 
\begin{tabular}{c|ccccccccc|c}
\hline
\multirow{3}{*}{\begin{tabular}[c]{@{}c@{}}Task completion\\ time command {[}s{]}\end{tabular}} & \multicolumn{9}{c|}{Position command}                                                                                                                                                                                                                                                                                                                            & \multirow{3}{*}{\begin{tabular}[c]{@{}c@{}}Speed\\ total\end{tabular}} \\ \cline{2-10}
                                                                                                & \multicolumn{4}{c|}{Learned}                                                                                                                                         & \multicolumn{5}{c|}{Unlearned}                                                                                                                                                            &                                                                        \\ \cline{2-10}
                                                                                                & \multicolumn{1}{c|}{Lower left}        & \multicolumn{1}{c|}{Upper left}         & \multicolumn{1}{c|}{Lower right}        & \multicolumn{1}{c|}{Upper right}        & \multicolumn{1}{c|}{Center}             & \multicolumn{1}{c|}{Left}               & \multicolumn{1}{c|}{Lower}              & \multicolumn{1}{c|}{Right}             & Upper              &                                                                        \\ \hline
3.00                                                                                            & \multicolumn{1}{c|}{0.0(0/3)}          & \multicolumn{1}{c|}{\textbf{33.3(1/3)}} & \multicolumn{1}{c|}{\textbf{100(3/3)}}  & \multicolumn{1}{c|}{\textbf{66.7(2/3)}} & \multicolumn{1}{c|}{\textbf{33.3(1/3)}} & \multicolumn{1}{c|}{0.0(0/3)}           & \multicolumn{1}{c|}{\textbf{100(3/3)}}  & \multicolumn{1}{c|}{\textbf{100(3/3)}} & \textbf{66.7(2/3)} & 55.6(15/27)                                                            \\ \hline
4.00                                                                                            & \multicolumn{1}{c|}{\textbf{100(3/3)}} & \multicolumn{1}{c|}{\textbf{100(3/3)}}  & \multicolumn{1}{c|}{\textbf{100(3/3)}}  & \multicolumn{1}{c|}{\textbf{100(3/3)}}  & \multicolumn{1}{c|}{\textbf{100(3/3)}}  & \multicolumn{1}{c|}{\textbf{66.7(2/3)}} & \multicolumn{1}{c|}{\textbf{100(3/3)}}  & \multicolumn{1}{c|}{\textbf{100(3/3)}} & \textbf{100(3/3)}  & 96.3(26/27)                                                            \\ \hline
5.00                                                                                            & \multicolumn{1}{c|}{\textbf{100(3/3)}} & \multicolumn{1}{c|}{\textbf{66.7(2/3)}} & \multicolumn{1}{c|}{\textbf{66.7(2/3)}} & \multicolumn{1}{c|}{\textbf{100(3/3)}}  & \multicolumn{1}{c|}{\textbf{100(3/3)}}  & \multicolumn{1}{c|}{\textbf{33.3(1/3)}} & \multicolumn{1}{c|}{\textbf{100(3/3)}}  & \multicolumn{1}{c|}{\textbf{100(3/3)}} & 0.0(0/3)           & 74.1(20/27)                                                            \\ \hline
6.00                                                                                            & \multicolumn{1}{c|}{33.3(1/3)}         & \multicolumn{1}{c|}{\textbf{33.3(1/3)}} & \multicolumn{1}{c|}{\textbf{100(3/3)}}  & \multicolumn{1}{c|}{\textbf{100(3/3)}}  & \multicolumn{1}{c|}{\textbf{100(3/3)}}  & \multicolumn{1}{c|}{\textbf{100(3/3)}}  & \multicolumn{1}{c|}{\textbf{100(3/3)}}  & \multicolumn{1}{c|}{66.7(2/3)}         & 0.0(0/3)           & 70.4(19/27)                                                            \\ \hline
7.00                                                                                            & \multicolumn{1}{c|}{\textbf{100(3/3)}} & \multicolumn{1}{c|}{\textbf{100(3/3)}}  & \multicolumn{1}{c|}{\textbf{100(3/3)}}  & \multicolumn{1}{c|}{\textbf{100(3/3)}}  & \multicolumn{1}{c|}{\textbf{66.7(2/3)}} & \multicolumn{1}{c|}{\textbf{100(3/3)}}  & \multicolumn{1}{c|}{\textbf{100(3/3)}}  & \multicolumn{1}{c|}{\textbf{100(3/3)}} & 0.0(0/3)           & 85.2(23/27)                                                            \\ \hline
8.00                                                                                            & \multicolumn{1}{c|}{\textbf{100(3/3)}} & \multicolumn{1}{c|}{\textbf{100(3/3)}}  & \multicolumn{1}{c|}{\textbf{100(3/3)}}  & \multicolumn{1}{c|}{\textbf{100(3/3)}}  & \multicolumn{1}{c|}{\textbf{100(3/3)}}  & \multicolumn{1}{c|}{\textbf{100(3/3)}}  & \multicolumn{1}{c|}{\textbf{100(3/3)}}  & \multicolumn{1}{c|}{\textbf{100(3/3)}} & \textbf{100(3/3)}  & 100(27/27)                                                             \\ \hline
9.00                                                                                            & \multicolumn{1}{c|}{\textbf{100(3/3)}} & \multicolumn{1}{c|}{\textbf{100(3/3)}}  & \multicolumn{1}{c|}{\textbf{100(3/3)}}  & \multicolumn{1}{c|}{\textbf{100(3/3)}}  & \multicolumn{1}{c|}{33.3(1/3)}          & \multicolumn{1}{c|}{\textbf{100(3/3)}}  & \multicolumn{1}{c|}{\textbf{100(3/3)}}  & \multicolumn{1}{c|}{\textbf{100(3/3)}} & \textbf{100(3/3)}  & 92.6(25/27)                                                            \\ \hline
10.0                                                                                            & \multicolumn{1}{c|}{\textbf{100(3/3)}} & \multicolumn{1}{c|}{\textbf{100(3/3)}}  & \multicolumn{1}{c|}{\textbf{100(3/3)}}  & \multicolumn{1}{c|}{\textbf{100(3/3)}}  & \multicolumn{1}{c|}{\textbf{100(3/3)}}  & \multicolumn{1}{c|}{\textbf{100(3/3)}}  & \multicolumn{1}{c|}{\textbf{66.7(2/3)}} & \multicolumn{1}{c|}{\textbf{100(3/3)}} & \textbf{100(3/3)}  & 96.3(26/27)                                                            \\ \hline
11.0                                                                                            & \multicolumn{1}{c|}{\textbf{100(3/3)}} & \multicolumn{1}{c|}{\textbf{100(3/3)}}  & \multicolumn{1}{c|}{\textbf{100(3/3)}}  & \multicolumn{1}{c|}{\textbf{100(3/3)}}  & \multicolumn{1}{c|}{\textbf{66.7(2/3)}} & \multicolumn{1}{c|}{33.3(1/3)}          & \multicolumn{1}{c|}{\textbf{0.0(0/3)}}  & \multicolumn{1}{c|}{\textbf{100(3/3)}} & \textbf{100(3/3)}  & 77.8(21/27)                                                            \\ \hline
12.0                                                                                            & \multicolumn{1}{c|}{\textbf{100(3/3)}} & \multicolumn{1}{c|}{\textbf{100(3/3)}}  & \multicolumn{1}{c|}{\textbf{100(3/3)}}  & \multicolumn{1}{c|}{\textbf{100(3/3)}}  & \multicolumn{1}{c|}{\textbf{100(3/3)}}  & \multicolumn{1}{c|}{\textbf{100(3/3)}}  & \multicolumn{1}{c|}{\textbf{100(3/3)}}  & \multicolumn{1}{c|}{\textbf{100(3/3)}} & \textbf{66.7(2/3)} & 96.3(26/27)                                                            \\ \hline
13.0                                                                                            & \multicolumn{1}{c|}{0.0(0/3)}          & \multicolumn{1}{c|}{\textbf{100(3/3)}}  & \multicolumn{1}{c|}{\textbf{66.7(2/3)}} & \multicolumn{1}{c|}{\textbf{100(3/3)}}  & \multicolumn{1}{c|}{\textbf{100(3/3)}}  & \multicolumn{1}{c|}{\textbf{100(3/3)}}  & \multicolumn{1}{c|}{\textbf{33.3(1/3)}} & \multicolumn{1}{c|}{\textbf{100(3/3)}} & \textbf{33.3(1/3)} & 70.4(19/27)                                                            \\ \hline
\begin{tabular}[c]{@{}c@{}}Position \\ total\end{tabular}                                       & \multicolumn{1}{c|}{75.8(25/33)}       & \multicolumn{1}{c|}{84.8(28/33)}        & \multicolumn{1}{c|}{93.9(31/33)}        & \multicolumn{1}{c|}{97.0(32/33)}        & \multicolumn{1}{c|}{81.8(27/33)}        & \multicolumn{1}{c|}{75.8(25/33)}        & \multicolumn{1}{c|}{81.8(27/33)}        & \multicolumn{1}{c|}{97.0(32/33)}       & 60.6(20/33)        & 83.2(247/297)                                                          \\ \hline
\end{tabular}
}
\label{tab:tab9}
\end{table*}

\begin{table*}[]
\caption{Task success rate of the self-supervised learning model 1 for pancake of 90g}
\scalebox{0.88}{ 
\begin{tabular}{c|ccccccccc|c}
\hline
\multirow{3}{*}{\begin{tabular}[c]{@{}c@{}}Task completion\\ time command {[}s{]}\end{tabular}} & \multicolumn{9}{c|}{Position command}                                                                                                                                                                                                                                                                                                                      & \multirow{3}{*}{\begin{tabular}[c]{@{}c@{}}Speed\\ total\end{tabular}} \\ \cline{2-10}
                                                                                                & \multicolumn{4}{c|}{Learned}                                                                                                                                       & \multicolumn{5}{c|}{Unlearned}                                                                                                                                                        &                                                                        \\ \cline{2-10}
                                                                                                & \multicolumn{1}{c|}{Lower left}        & \multicolumn{1}{c|}{Upper left}         & \multicolumn{1}{c|}{Lower right}       & \multicolumn{1}{c|}{Upper right}       & \multicolumn{1}{c|}{Center}            & \multicolumn{1}{c|}{Left}              & \multicolumn{1}{c|}{Lower}             & \multicolumn{1}{c|}{Right}             & Upper             &                                                                        \\ \hline
3.00                                                                                            & \multicolumn{1}{c|}{0.0(0/3)}          & \multicolumn{1}{c|}{\textbf{66.7(2/3)}} & \multicolumn{1}{c|}{\textbf{100(3/3)}} & \multicolumn{1}{c|}{\textbf{100(3/3)}} & \multicolumn{1}{c|}{66.7(2/3)}         & \multicolumn{1}{c|}{0.0(0/3)}          & \multicolumn{1}{c|}{\textbf{100(3/3)}} & \multicolumn{1}{c|}{\textbf{100(3/3)}} & 66.7(2/3)         & 66.7(18/27)                                                            \\ \hline
4.00                                                                                            & \multicolumn{1}{c|}{\textbf{100(3/3)}} & \multicolumn{1}{c|}{\textbf{33.3(1/3)}} & \multicolumn{1}{c|}{\textbf{100(3/3)}} & \multicolumn{1}{c|}{\textbf{100(3/3)}} & \multicolumn{1}{c|}{\textbf{100(3/3)}} & \multicolumn{1}{c|}{\textbf{0.0(0/3)}} & \multicolumn{1}{c|}{\textbf{100(3/3)}} & \multicolumn{1}{c|}{\textbf{100(3/3)}} & \textbf{100(3/3)} & 81.5(22/27)                                                            \\ \hline
5.00                                                                                            & \multicolumn{1}{c|}{\textbf{100(3/3)}} & \multicolumn{1}{c|}{\textbf{66.7(2/3)}} & \multicolumn{1}{c|}{\textbf{100(3/3)}} & \multicolumn{1}{c|}{\textbf{100(3/3)}} & \multicolumn{1}{c|}{\textbf{100(3/3)}} & \multicolumn{1}{c|}{\textbf{0.0(0/3)}} & \multicolumn{1}{c|}{\textbf{100(3/3)}} & \multicolumn{1}{c|}{66.7(2/3)}         & 0.0(0/3)          & 70.4(19/27)                                                            \\ \hline
6.00                                                                                            & \multicolumn{1}{c|}{66.7(2/3)}         & \multicolumn{1}{c|}{0.0(0/3)}           & \multicolumn{1}{c|}{\textbf{100(3/3)}} & \multicolumn{1}{c|}{\textbf{100(3/3)}} & \multicolumn{1}{c|}{\textbf{100(3/3)}} & \multicolumn{1}{c|}{\textbf{100(3/3)}} & \multicolumn{1}{c|}{\textbf{100(3/3)}} & \multicolumn{1}{c|}{0.0(0/3)}          & 0.0(0/3)          & 63.0(17/27)                                                            \\ \hline
7.00                                                                                            & \multicolumn{1}{c|}{\textbf{100(3/3)}} & \multicolumn{1}{c|}{\textbf{100(3/3)}}  & \multicolumn{1}{c|}{\textbf{100(3/3)}} & \multicolumn{1}{c|}{\textbf{100(3/3)}} & \multicolumn{1}{c|}{66.7(2/3)}         & \multicolumn{1}{c|}{\textbf{100(3/3)}} & \multicolumn{1}{c|}{\textbf{100(3/3)}} & \multicolumn{1}{c|}{\textbf{100(3/3)}} & 33.3(1/3)         & 88.9(24/27)                                                            \\ \hline
8.00                                                                                            & \multicolumn{1}{c|}{\textbf{100(3/3)}} & \multicolumn{1}{c|}{\textbf{100(3/3)}}  & \multicolumn{1}{c|}{\textbf{100(3/3)}} & \multicolumn{1}{c|}{\textbf{100(3/3)}} & \multicolumn{1}{c|}{66.7(2/3)}         & \multicolumn{1}{c|}{\textbf{100(3/3)}} & \multicolumn{1}{c|}{\textbf{100(3/3)}} & \multicolumn{1}{c|}{\textbf{100(3/3)}} & 66.7(2/3)         & 92.6(25/27)                                                            \\ \hline
9.00                                                                                            & \multicolumn{1}{c|}{\textbf{100(3/3)}} & \multicolumn{1}{c|}{\textbf{100(3/3)}}  & \multicolumn{1}{c|}{\textbf{100(3/3)}} & \multicolumn{1}{c|}{\textbf{100(3/3)}} & \multicolumn{1}{c|}{33.3(1/3)}         & \multicolumn{1}{c|}{\textbf{100(3/3)}} & \multicolumn{1}{c|}{\textbf{100(3/3)}} & \multicolumn{1}{c|}{\textbf{100(3/3)}} & \textbf{100(3/3)} & 92.6(25/27)                                                            \\ \hline
10.0                                                                                            & \multicolumn{1}{c|}{\textbf{100(3/3)}} & \multicolumn{1}{c|}{\textbf{100(3/3)}}  & \multicolumn{1}{c|}{\textbf{100(3/3)}} & \multicolumn{1}{c|}{\textbf{100(3/3)}} & \multicolumn{1}{c|}{\textbf{100(3/3)}} & \multicolumn{1}{c|}{\textbf{100(3/3)}} & \multicolumn{1}{c|}{33.3(1/3)}         & \multicolumn{1}{c|}{\textbf{100(3/3)}} & \textbf{100(3/3)} & 92.6(25/27)                                                            \\ \hline
11.0                                                                                            & \multicolumn{1}{c|}{\textbf{100(3/3)}} & \multicolumn{1}{c|}{\textbf{100(3/3)}}  & \multicolumn{1}{c|}{\textbf{100(3/3)}} & \multicolumn{1}{c|}{\textbf{100(3/3)}} & \multicolumn{1}{c|}{66.7(2/3)}         & \multicolumn{1}{c|}{66.7(2/3)}         & \multicolumn{1}{c|}{0.0(0/3)}          & \multicolumn{1}{c|}{\textbf{100(3/3)}} & 66.7(2/3)         & 77.8(21/27)                                                            \\ \hline
12.0                                                                                            & \multicolumn{1}{c|}{\textbf{100(3/3)}} & \multicolumn{1}{c|}{\textbf{100(3/3)}}  & \multicolumn{1}{c|}{\textbf{100(3/3)}} & \multicolumn{1}{c|}{\textbf{100(3/3)}} & \multicolumn{1}{c|}{\textbf{100(3/3)}} & \multicolumn{1}{c|}{\textbf{100(3/3)}} & \multicolumn{1}{c|}{\textbf{100(3/3)}} & \multicolumn{1}{c|}{\textbf{100(3/3)}} & 0.0(0/3)          & 88.9(24/27)                                                            \\ \hline
13.0                                                                                            & \multicolumn{1}{c|}{0.0(0/3)}          & \multicolumn{1}{c|}{\textbf{66.7(2/3)}} & \multicolumn{1}{c|}{33.3(1/3)}         & \multicolumn{1}{c|}{\textbf{100(3/3)}} & \multicolumn{1}{c|}{\textbf{100(3/3)}} & \multicolumn{1}{c|}{\textbf{100(3/3)}} & \multicolumn{1}{c|}{66.7(2/3)}         & \multicolumn{1}{c|}{\textbf{100(3/3)}} & 33.3(1/3)         & 66.7(18/27)                                                            \\ \hline
\begin{tabular}[c]{@{}c@{}}Position \\ total\end{tabular}                                       & \multicolumn{1}{c|}{78.8(26/33)}       & \multicolumn{1}{c|}{75.8(25/33)}        & \multicolumn{1}{c|}{93.9(31/33)}       & \multicolumn{1}{c|}{100(33/33)}        & \multicolumn{1}{c|}{81.8(27/33)}       & \multicolumn{1}{c|}{69.7(23/33)}       & \multicolumn{1}{c|}{81.8(27/33)}       & \multicolumn{1}{c|}{87.9(29/33)}       & 51.5(17/33)       & 80.1(238/297)                                                          \\ \hline
\end{tabular}
}
\label{tab:tab10}
\end{table*}

\begin{table*}[]
\caption{Task success rate of the self-supervised learning model 2 for pancake of 30g}
\scalebox{0.88}{ 
\begin{tabular}{c|ccccccccc|c}
\hline
\multirow{3}{*}{\begin{tabular}[c]{@{}c@{}}Task completion\\ time command {[}s{]}\end{tabular}} & \multicolumn{9}{c|}{Position command}                                                                                                                                                                                                                                                                                                                          & \multirow{3}{*}{\begin{tabular}[c]{@{}c@{}}Speed\\ total\end{tabular}} \\ \cline{2-10}
                                                                                                & \multicolumn{4}{c|}{Learned}                                                                                                                                       & \multicolumn{5}{c|}{Unlearned}                                                                                                                                                            &                                                                        \\ \cline{2-10}
                                                                                                & \multicolumn{1}{c|}{Lower left}        & \multicolumn{1}{c|}{Upper left}        & \multicolumn{1}{c|}{Lower right}       & \multicolumn{1}{c|}{Upper right}        & \multicolumn{1}{c|}{Center}             & \multicolumn{1}{c|}{Left}               & \multicolumn{1}{c|}{Lower}              & \multicolumn{1}{c|}{Right}             & Upper              &                                                                        \\ \hline
3.00                                                                                            & \multicolumn{1}{c|}{66.7(2/3)}         & \multicolumn{1}{c|}{\textbf{100(3/3)}} & \multicolumn{1}{c|}{\textbf{100(3/3)}} & \multicolumn{1}{c|}{\textbf{100(3/3)}}  & \multicolumn{1}{c|}{66.7(2/3)}          & \multicolumn{1}{c|}{66.7(2/3)}          & \multicolumn{1}{c|}{\textbf{100(3/3)}}  & \multicolumn{1}{c|}{66.7(2/3)}         & \textbf{100(3/3)}  & 85.2(23/27)                                                            \\ \hline
4.00                                                                                            & \multicolumn{1}{c|}{\textbf{100(3/3)}} & \multicolumn{1}{c|}{\textbf{100(3/3)}} & \multicolumn{1}{c|}{\textbf{100(3/3)}} & \multicolumn{1}{c|}{\textbf{100(3/3)}}  & \multicolumn{1}{c|}{\textbf{66.7(2/3)}} & \multicolumn{1}{c|}{\textbf{66.7(2/3)}} & \multicolumn{1}{c|}{\textbf{100(3/3)}}  & \multicolumn{1}{c|}{\textbf{100(3/3)}} & \textbf{100(3/3)}  & 92.6(25/27)                                                            \\ \hline
5.00                                                                                            & \multicolumn{1}{c|}{66.7(2/3)}         & \multicolumn{1}{c|}{\textbf{100(3/3)}} & \multicolumn{1}{c|}{0.0(0/3)}          & \multicolumn{1}{c|}{\textbf{33.3(1/3)}} & \multicolumn{1}{c|}{\textbf{66.7(2/3)}} & \multicolumn{1}{c|}{\textbf{100(3/3)}}  & \multicolumn{1}{c|}{\textbf{100(3/3)}}  & \multicolumn{1}{c|}{\textbf{100(3/3)}} & 66.7(2/3)          & 70.4(19/27)                                                            \\ \hline
6.00                                                                                            & \multicolumn{1}{c|}{33.3(1/3)}         & \multicolumn{1}{c|}{66.7(2/3)}         & \multicolumn{1}{c|}{33.3(1/3)}         & \multicolumn{1}{c|}{\textbf{100(3/3)}}  & \multicolumn{1}{c|}{\textbf{33.3(1/3)}} & \multicolumn{1}{c|}{\textbf{100(3/3)}}  & \multicolumn{1}{c|}{\textbf{33.3(1/3)}} & \multicolumn{1}{c|}{66.7(2/3)}         & \textbf{33.3(1/3)} & 55.6(15/27)                                                            \\ \hline
7.00                                                                                            & \multicolumn{1}{c|}{66.7(2/3)}         & \multicolumn{1}{c|}{\textbf{100(3/3)}} & \multicolumn{1}{c|}{\textbf{100(3/3)}} & \multicolumn{1}{c|}{\textbf{100(3/3)}}  & \multicolumn{1}{c|}{\textbf{100(3/3)}}  & \multicolumn{1}{c|}{\textbf{100(3/3)}}  & \multicolumn{1}{c|}{\textbf{100(3/3)}}  & \multicolumn{1}{c|}{\textbf{100(3/3)}} & \textbf{100(3/3)}  & 96.3(26/27)                                                            \\ \hline
8.00                                                                                            & \multicolumn{1}{c|}{\textbf{100(3/3)}} & \multicolumn{1}{c|}{\textbf{100(3/3)}} & \multicolumn{1}{c|}{\textbf{100(3/3)}} & \multicolumn{1}{c|}{\textbf{100(3/3)}}  & \multicolumn{1}{c|}{\textbf{100(3/3)}}  & \multicolumn{1}{c|}{\textbf{100(3/3)}}  & \multicolumn{1}{c|}{\textbf{100(3/3)}}  & \multicolumn{1}{c|}{0.0(0/3)}          & 66.7(2/3)          & 85.2(23/27)                                                            \\ \hline
9.00                                                                                            & \multicolumn{1}{c|}{\textbf{100(3/3)}} & \multicolumn{1}{c|}{66.7(2/3)}         & \multicolumn{1}{c|}{\textbf{100(3/3)}} & \multicolumn{1}{c|}{\textbf{100(3/3)}}  & \multicolumn{1}{c|}{\textbf{33.3(1/3)}} & \multicolumn{1}{c|}{\textbf{100(3/3)}}  & \multicolumn{1}{c|}{\textbf{100(3/3)}}  & \multicolumn{1}{c|}{\textbf{100(3/3)}} & \textbf{100(3/3)}  & 88.9(24/27)                                                            \\ \hline
10.0                                                                                            & \multicolumn{1}{c|}{\textbf{100(3/3)}} & \multicolumn{1}{c|}{\textbf{100(3/3)}} & \multicolumn{1}{c|}{\textbf{100(3/3)}} & \multicolumn{1}{c|}{\textbf{100(3/3)}}  & \multicolumn{1}{c|}{\textbf{100(3/3)}}  & \multicolumn{1}{c|}{\textbf{100(3/3)}}  & \multicolumn{1}{c|}{66.7(2/3)}          & \multicolumn{1}{c|}{\textbf{100(3/3)}} & \textbf{100(3/3)}  & 96.3(26/27)                                                            \\ \hline
11.0                                                                                            & \multicolumn{1}{c|}{\textbf{100(3/3)}} & \multicolumn{1}{c|}{\textbf{100(3/3)}} & \multicolumn{1}{c|}{66.7(2/3)}         & \multicolumn{1}{c|}{\textbf{100(3/3)}}  & \multicolumn{1}{c|}{\textbf{100(3/3)}}  & \multicolumn{1}{c|}{\textbf{100(3/3)}}  & \multicolumn{1}{c|}{66.7(2/3)}          & \multicolumn{1}{c|}{\textbf{100(3/3)}} & \textbf{100(3/3)}  & 92.6(25/27)                                                            \\ \hline
12.0                                                                                            & \multicolumn{1}{c|}{66.7(2/3)}         & \multicolumn{1}{c|}{\textbf{100(3/3)}} & \multicolumn{1}{c|}{66.7(2/3)}         & \multicolumn{1}{c|}{\textbf{100(3/3)}}  & \multicolumn{1}{c|}{\textbf{100(3/3)}}  & \multicolumn{1}{c|}{\textbf{100(3/3)}}  & \multicolumn{1}{c|}{\textbf{100(3/3)}}  & \multicolumn{1}{c|}{\textbf{100(3/3)}} & 66.7(2/3)          & 88.9(24/27)                                                            \\ \hline
13.0                                                                                            & \multicolumn{1}{c|}{66.7(2/3)}         & \multicolumn{1}{c|}{\textbf{100(3/3)}} & \multicolumn{1}{c|}{\textbf{100(3/3)}} & \multicolumn{1}{c|}{\textbf{100(3/3)}}  & \multicolumn{1}{c|}{\textbf{100(3/3)}}  & \multicolumn{1}{c|}{\textbf{100(3/3)}}  & \multicolumn{1}{c|}{66.7(2/3)}          & \multicolumn{1}{c|}{\textbf{100(3/3)}} & 66.7(2/3)          & 88.9(24/27)                                                            \\ \hline
\begin{tabular}[c]{@{}c@{}}Position \\ total\end{tabular}                                       & \multicolumn{1}{c|}{78.8(26/33)}       & \multicolumn{1}{c|}{93.9(31/33)}       & \multicolumn{1}{c|}{78.8(26/33)}       & \multicolumn{1}{c|}{93.9(31/33)}        & \multicolumn{1}{c|}{78.8(26/33)}        & \multicolumn{1}{c|}{93.9(31/33)}        & \multicolumn{1}{c|}{84.8(28/33)}        & \multicolumn{1}{c|}{84.8(28/33)}       & 81.8(27/33)        & 85.5(254/297)                                                          \\ \hline
\end{tabular}
}
\label{tab:tab11}
\end{table*}

\begin{table*}[]
\caption{Task success rate of the self-supervised learning model 2 for pancake of 90g}
\scalebox{0.88}{ 
\begin{tabular}{c|ccccccccc|c}
\hline
\multirow{3}{*}{\begin{tabular}[c]{@{}c@{}}Task completion\\ time command {[}s{]}\end{tabular}} & \multicolumn{9}{c|}{Position command}                                                                                                                                                                                                                                                                                                                           & \multirow{3}{*}{\begin{tabular}[c]{@{}c@{}}Speed\\ total\end{tabular}} \\ \cline{2-10}
                                                                                                & \multicolumn{4}{c|}{Learned}                                                                                                                                       & \multicolumn{5}{c|}{Unlearned}                                                                                                                                                             &                                                                        \\ \cline{2-10}
                                                                                                & \multicolumn{1}{c|}{Lower left}        & \multicolumn{1}{c|}{Upper left}        & \multicolumn{1}{c|}{Lower right}        & \multicolumn{1}{c|}{Upper right}       & \multicolumn{1}{c|}{Center}             & \multicolumn{1}{c|}{Left}               & \multicolumn{1}{c|}{Lower}              & \multicolumn{1}{c|}{Right}              & Upper              &                                                                        \\ \hline
3.00                                                                                            & \multicolumn{1}{c|}{66.7(2/3)}         & \multicolumn{1}{c|}{\textbf{100(3/3)}} & \multicolumn{1}{c|}{\textbf{66.7(2/3)}} & \multicolumn{1}{c|}{\textbf{100(3/3)}} & \multicolumn{1}{c|}{\textbf{100(3/3)}}  & \multicolumn{1}{c|}{0.0(0/3)}           & \multicolumn{1}{c|}{\textbf{66.7(2/3)}} & \multicolumn{1}{c|}{\textbf{33.3(1/3)}} & \textbf{100(3/3)}  & 70.4(19/27)                                                            \\ \hline
4.00                                                                                            & \multicolumn{1}{c|}{\textbf{100(3/3)}} & \multicolumn{1}{c|}{\textbf{100(3/3)}} & \multicolumn{1}{c|}{\textbf{100(3/3)}}  & \multicolumn{1}{c|}{\textbf{100(3/3)}} & \multicolumn{1}{c|}{\textbf{100(3/3)}}  & \multicolumn{1}{c|}{\textbf{66.7(2/3)}} & \multicolumn{1}{c|}{\textbf{66.7(2/3)}} & \multicolumn{1}{c|}{\textbf{100(3/3)}}  & \textbf{100(3/3)}  & 92.6(25/27)                                                            \\ \hline
5.00                                                                                            & \multicolumn{1}{c|}{0.0(0/3)}          & \multicolumn{1}{c|}{\textbf{100(3/3)}} & \multicolumn{1}{c|}{0.0(0/3)}           & \multicolumn{1}{c|}{\textbf{100(3/3)}} & \multicolumn{1}{c|}{\textbf{100(3/3)}}  & \multicolumn{1}{c|}{\textbf{100(3/3)}}  & \multicolumn{1}{c|}{\textbf{100(3/3)}}  & \multicolumn{1}{c|}{\textbf{100(3/3)}}  & \textbf{100(3/3)}  & 77.8(21/27)                                                            \\ \hline
6.00                                                                                            & \multicolumn{1}{c|}{\textbf{100(3/3)}} & \multicolumn{1}{c|}{\textbf{100(3/3)}} & \multicolumn{1}{c|}{\textbf{100(3/3)}}  & \multicolumn{1}{c|}{\textbf{100(3/3)}} & \multicolumn{1}{c|}{\textbf{66.7(2/3)}} & \multicolumn{1}{c|}{\textbf{100(3/3)}}  & \multicolumn{1}{c|}{\textbf{0.0(0/3)}}  & \multicolumn{1}{c|}{\textbf{100(3/3)}}  & \textbf{0.0(0/3)}  & 74.1(20/27)                                                            \\ \hline
7.00                                                                                            & \multicolumn{1}{c|}{66.7(2/3)}         & \multicolumn{1}{c|}{\textbf{100(3/3)}} & \multicolumn{1}{c|}{\textbf{100(3/3)}}  & \multicolumn{1}{c|}{\textbf{100(3/3)}} & \multicolumn{1}{c|}{\textbf{100(3/3)}}  & \multicolumn{1}{c|}{\textbf{100(3/3)}}  & \multicolumn{1}{c|}{\textbf{66.7(2/3)}} & \multicolumn{1}{c|}{\textbf{100(3/3)}}  & \textbf{100(3/3)}  & 92.6(25/27)                                                            \\ \hline
8.00                                                                                            & \multicolumn{1}{c|}{\textbf{100(3/3)}} & \multicolumn{1}{c|}{\textbf{100(3/3)}} & \multicolumn{1}{c|}{\textbf{100(3/3)}}  & \multicolumn{1}{c|}{\textbf{100(3/3)}} & \multicolumn{1}{c|}{\textbf{66.7(2/3)}} & \multicolumn{1}{c|}{\textbf{100(3/3)}}  & \multicolumn{1}{c|}{\textbf{100(3/3)}}  & \multicolumn{1}{c|}{66.7(2/3)}          & 66.7(2/3)          & 88.9(24/27)                                                            \\ \hline
9.00                                                                                            & \multicolumn{1}{c|}{\textbf{100(3/3)}} & \multicolumn{1}{c|}{\textbf{100(3/3)}} & \multicolumn{1}{c|}{\textbf{100(3/3)}}  & \multicolumn{1}{c|}{\textbf{100(3/3)}} & \multicolumn{1}{c|}{\textbf{100(3/3)}}  & \multicolumn{1}{c|}{\textbf{100(3/3)}}  & \multicolumn{1}{c|}{\textbf{100(3/3)}}  & \multicolumn{1}{c|}{\textbf{100(3/3)}}  & \textbf{100(3/3)}  & 100(27/27)                                                             \\ \hline
10.0                                                                                            & \multicolumn{1}{c|}{\textbf{100(3/3)}} & \multicolumn{1}{c|}{\textbf{100(3/3)}} & \multicolumn{1}{c|}{\textbf{66.7(2/3)}} & \multicolumn{1}{c|}{\textbf{100(3/3)}} & \multicolumn{1}{c|}{\textbf{100(3/3)}}  & \multicolumn{1}{c|}{\textbf{66.7(2/3)}} & \multicolumn{1}{c|}{66.7(2/3)}          & \multicolumn{1}{c|}{\textbf{100(3/3)}}  & \textbf{100(3/3)}  & 88.9(24/27)                                                            \\ \hline
11.0                                                                                            & \multicolumn{1}{c|}{\textbf{100(3/3)}} & \multicolumn{1}{c|}{\textbf{100(3/3)}} & \multicolumn{1}{c|}{0.0(0/3)}           & \multicolumn{1}{c|}{\textbf{100(3/3)}} & \multicolumn{1}{c|}{\textbf{100(3/3)}}  & \multicolumn{1}{c|}{\textbf{100(3/3)}}  & \multicolumn{1}{c|}{\textbf{100(3/3)}}  & \multicolumn{1}{c|}{\textbf{100(3/3)}}  & \textbf{100(3/3)}  & 88.9(24/27)                                                            \\ \hline
12.0                                                                                            & \multicolumn{1}{c|}{\textbf{100(3/3)}} & \multicolumn{1}{c|}{\textbf{100(3/3)}} & \multicolumn{1}{c|}{\textbf{100(3/3)}}  & \multicolumn{1}{c|}{\textbf{100(3/3)}} & \multicolumn{1}{c|}{\textbf{100(3/3)}}  & \multicolumn{1}{c|}{\textbf{100(3/3)}}  & \multicolumn{1}{c|}{\textbf{100(3/3)}}  & \multicolumn{1}{c|}{\textbf{100(3/3)}}  & 66.7(2/3)          & 96.3(26/27)                                                            \\ \hline
13.0                                                                                            & \multicolumn{1}{c|}{0.0(0/3)}          & \multicolumn{1}{c|}{\textbf{100(3/3)}} & \multicolumn{1}{c|}{\textbf{100(3/3)}}  & \multicolumn{1}{c|}{\textbf{100(3/3)}} & \multicolumn{1}{c|}{\textbf{100(3/3)}}  & \multicolumn{1}{c|}{\textbf{100(3/3)}}  & \multicolumn{1}{c|}{33.3(1/3)}          & \multicolumn{1}{c|}{\textbf{100(3/3)}}  & \textbf{33.3(1/3)} & 74.1(20/27)                                                            \\ \hline
\begin{tabular}[c]{@{}c@{}}Position \\ total\end{tabular}                                       & \multicolumn{1}{c|}{75.8(25/33)}       & \multicolumn{1}{c|}{100(33/33)}        & \multicolumn{1}{c|}{75.8(25/33)}        & \multicolumn{1}{c|}{100(33/33)}        & \multicolumn{1}{c|}{93.9(31/33)}        & \multicolumn{1}{c|}{84.8(28/33)}        & \multicolumn{1}{c|}{72.7(24/33)}        & \multicolumn{1}{c|}{90.9(30/33)}        & 78.8(26/33)        & 85.9(255/297)                                                          \\ \hline
\end{tabular}
}
\label{tab:tab12}
\end{table*}

\begin{IEEEbiography}[{\includegraphics[width=1.05in,height=1.30in,clip,keepaspectratio]{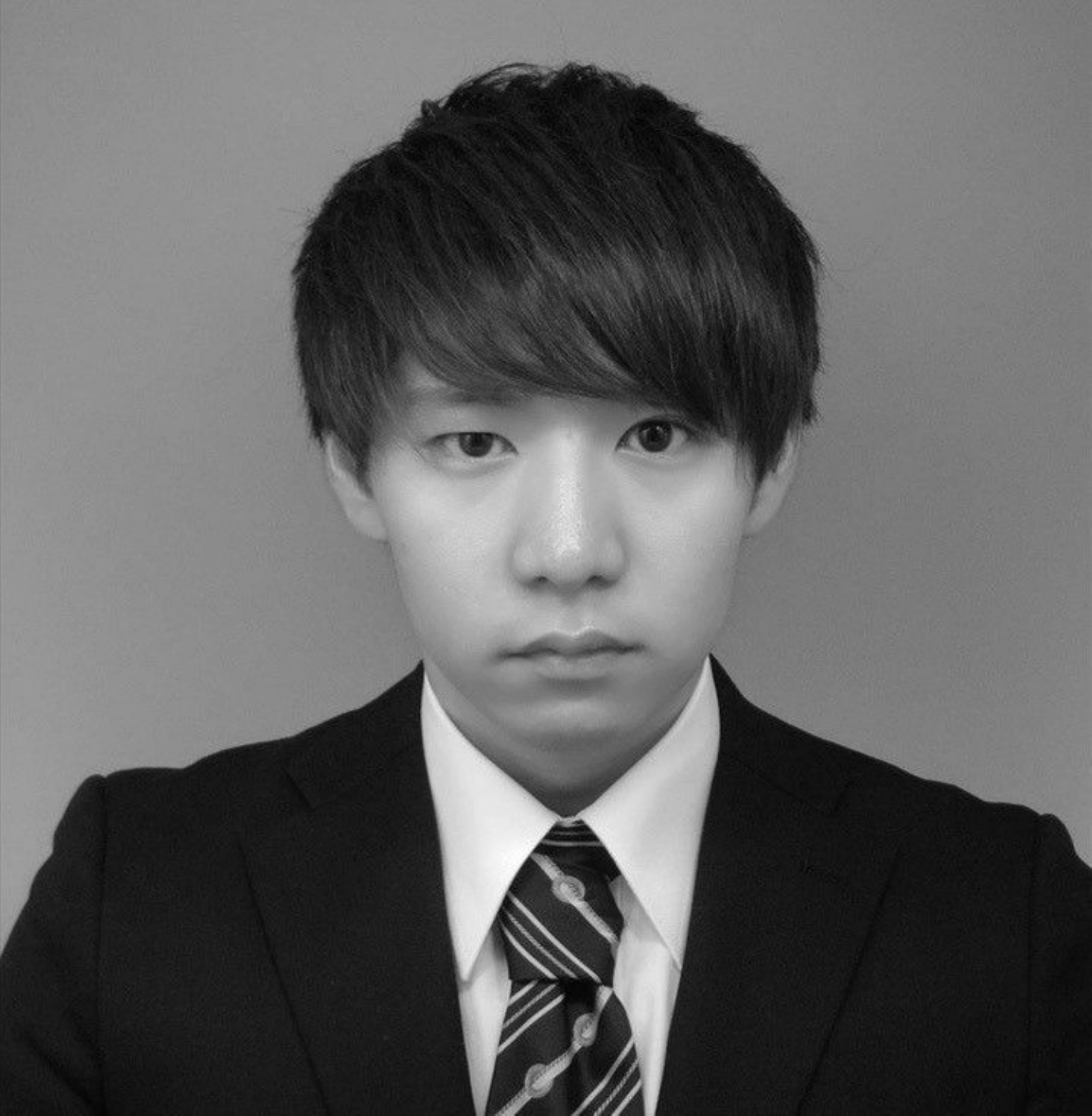}}]{Yuki Saigusa} received the B.E. degree in electrical and electronic system engineering from Saitama University, Saitama, Japan, in 2020. He is currently working toward M.E. degrees in the Graduate School of Science and Technology, and degree programs in intelligent and mechanical interaction systems at University of Tsukuba, Japan. His research interests include motion control, robotics, and machine learning. 
\end{IEEEbiography}

\begin{IEEEbiography}[{\includegraphics[width=1.05in,height=1.30in,clip,keepaspectratio]{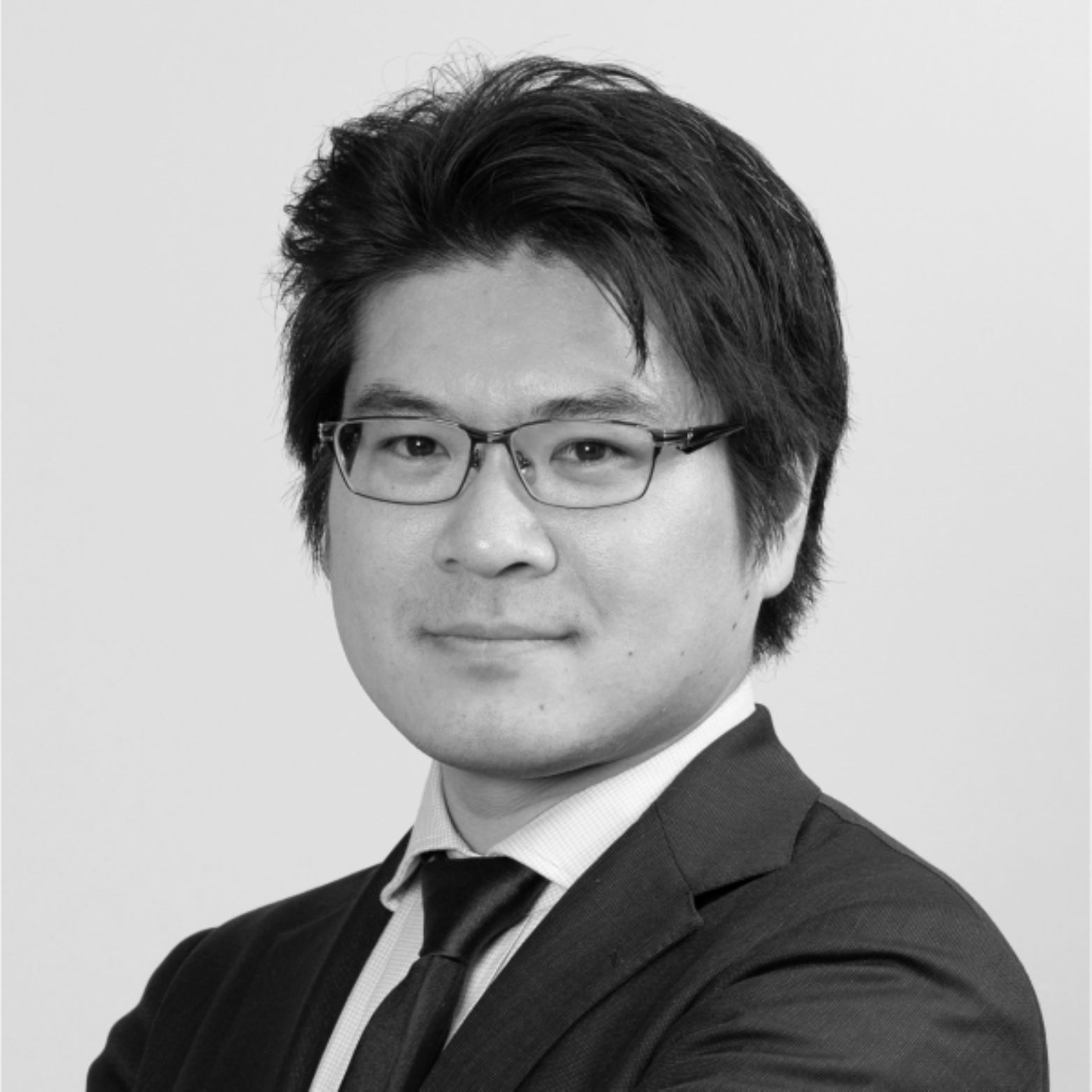}}]{Sho Sakaino} received the B.E. degree in system design engineering and the M.E. and Ph.D. degrees in integrated design engineering from Keio University, Yokohama, Japan, in 2006, 2008, and 2011, respectively. 
He was an assistant professor at Saitama University from 2011 to 2019. Since 2019, he has been an associate professor at University of Tsukuba. His research interests include mechatronics, motion control, robotics, and haptics. 
He received the IEEJ Industry Application Society Distinguished Transaction Paper Award in 2011 and 2020. He also received the RSJ Advanced Robotics Excellent Paper Award in 2020.
\end{IEEEbiography}

\begin{IEEEbiography}[{\includegraphics[width=1.05in,height=1.30in,clip,keepaspectratio]{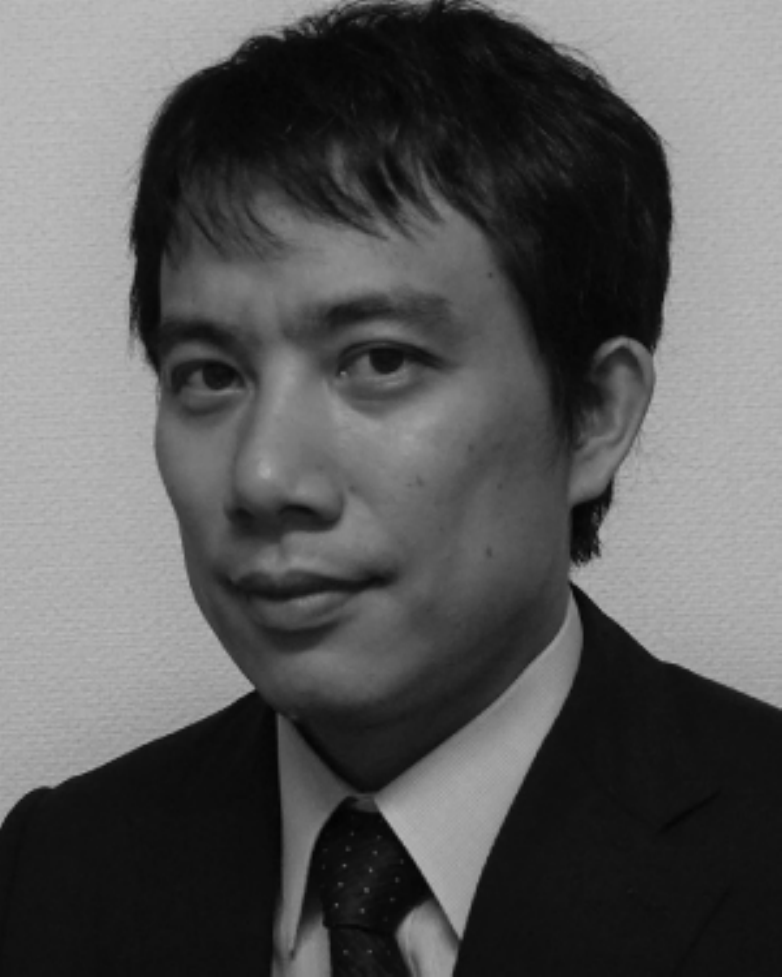}}]{Toshiaki Tsuji} received the B.E. degree in system design engineering and the M.E. and Ph.D. degrees in integrated design engineering from Keio University, Yokohama, Japan, in 2001, 2003, and 2006, respectively. 
He was a Research Associate in the Department of Mechanical Engineering, Tokyo University of Science, from 2006 to 2007. He is currently an Associate Professor in the Department of Electrical and Electronic Systems, Saitama University, Saitama, Japan. His research interests include motion control, haptics, and rehabilitation robots. 
Dr. Tsuji received the FANUC FA and Robot Foundation Original Paper Award in 2007 and 2008. He also received the RSJ Advanced Robotics Excellent Paper Award and the IEEJ Industry Application Society Distinguished Transaction Paper Award in 2020.
\end{IEEEbiography}

\EOD

\end{document}